\documentclass{article}

\PassOptionsToPackage{dvipsnames,table}{xcolor}

\usepackage[preprint]{neurips_2026}

\usepackage{float}          
\usepackage{graphicx}       

\usepackage{mathtools}
\usepackage{amsthm}
\usepackage{amssymb}
\usepackage{amsmath}
\usepackage{bm}
\usepackage{bbding}
\usepackage{caption}
\usepackage{subcaption}
\usepackage{enumitem}
\usepackage{dashrule}
\usepackage{titletoc}
\usepackage[dvipsnames,table]{xcolor}
\usepackage[most]{tcolorbox}
\tcbset{
  beigenote/.style={
    colback=panelA,
    colframe=panelA!80!black,
    boxrule=0.4pt,
    arc=2pt,
    left=6pt, right=6pt, top=4pt, bottom=4pt,
    fontupper=\normalsize,
  }
}
\newtcolorbox{beigebox}[1][]{beigenote,#1}

\definecolor{darkred}{rgb}{0.55, 0.0, 0.0}

\usepackage{rotating}
\usepackage[colorinlistoftodos,prependcaption,textsize=tiny]{todonotes}
\usepackage{booktabs}
\usepackage{wrapfig}
\usepackage{multirow}
\usepackage{tikz}
\usetikzlibrary{arrows.meta,positioning,calc,shapes.geometric,shapes.misc,decorations.pathreplacing,fit,backgrounds,shadows}
\usepackage{array}
\usepackage{hyperref}
\usepackage{cleveref} 
\Crefname{equation}{Eq.}{Eqs.}

\definecolor{mylink}{HTML}{1A2B3C}   
\definecolor{mycite}{HTML}{0F62FE}   
\definecolor{myurl}{HTML}{E83E8C}    
\definecolor{panelA}{HTML}{FFF6DC} 
\definecolor{panelB}{HTML}{EEF7F1} 
\definecolor{lightyellow}{HTML}{FFFFED}
\definecolor{lightpink}{HTML}{F8F1EF}
\definecolor{customBlue}{HTML}{BDD7EE}
\definecolor{tcwmOrange}{HTML}{D87F3C}
\definecolor{tcwmSoftOrange}{HTML}{E9A36F}
\definecolor{tcwmCream}{HTML}{FFF7EF}
\definecolor{tcwmBlue}{HTML}{0F4D92}
\newtcolorbox{tcwmtheorembox}[1]{
  enhanced,
  breakable,
  colback=white,
  colframe=black!50,
  colbacktitle=white,
  coltitle=black,
  title={#1},
  fonttitle=\sffamily\bfseries,
  titlerule=0.3pt,
  boxrule=0.4pt,
  arc=2pt,
  left=6pt, right=6pt, top=4pt, bottom=4pt,
  before skip=6pt,
  after skip=6pt,
}

\newcolumntype{A}{>{\columncolor{panelA}}c}
\newcolumntype{B}{>{\columncolor{panelB}}c}
\newcolumntype{G}{>{\columncolor{gray!25}}c}

\hypersetup{
  colorlinks=true,
  linkcolor=mycite,
  citecolor=gray,
  urlcolor=myurl
}

\newtheorem{definition}{Definition}

\def\xx{\mathbf{x}}
\def\zz{\mathbf{z}}
\def\ss{\mathbf{s}}

\def\oo{\mathbf{o}}
\def\aa{\mathbf{a}}

\def\hzz{\hat{\mathbf{z}}}

\newcommand{\mytitle}{\fontsize{14.8}{17.8}\selectfont Back to Parsimonious Latents: \\Learning Task-Centric World Models from Visual Foundations%
}

\newcommand{\method}{TC-WM}

\newcommand{\eg}{\emph{e.g.}}
\newcommand{\ie}{\emph{i.e.}}

\definecolor{nhred}{HTML}{D62727}

\title{\mytitle}

\author{%
  Minghao Fu$^{1}$ \quad Fan Feng$^{1}$ \quad Nicklas Hansen$^{1}$ \quad Biwei Huang$^{1}$ \\
  $^{1}$University of California, San Diego \\
  \texttt{\{m9fu, f2feng, nihansen, bih007\}@ucsd.edu}
}

\begin{document}

\maketitle

\begin{abstract}
World models enable agents to predict future dynamics conditioned on actions, making the choice of latent representation central to planning and control. 
Such representations are often either learned directly from pixels with limited semantic structure or inherited from frozen visual foundation models with excessive task-irrelevant detail, yielding state spaces that are poorly matched to downstream planning and control. This is especially challenging in reward-free offline settings, where the model must learn from fixed trajectories without reward supervision or online interaction.
To address this, we propose {\method}, a framework for turning foundation-model embeddings into compact, task-sufficient world representations. The key design is to treat the pretrained embedding space as a semantic scaffold rather than as the final state space: {\method} linearly projects high-dimensional visual embeddings into a compact latent as the dynamic space, aligns a subspace with the agent's physical state via contrastive learning, and reconstructs embeddings to preserve useful visual structure. This combines the generality of foundation features with the controllability of task-centric dynamics. Theoretically, we show that {\method} suffices to identify the underlying task-centric latent factors up to a simple transformation. Empirically, {\method} enables test-time planning across diverse environments (\eg, \texttt{Robomimic} and \texttt{D4RL}), achieving better world-modeling quality and more precise control than state-of-the-art approaches.
\begin{center}
    \vspace{-0.025in}
    \textbf{Webpage: \url{https://minghaofu.com/tc-wm/}}
\end{center}
\end{abstract}

\section{Introduction}
\label{sec:intro}

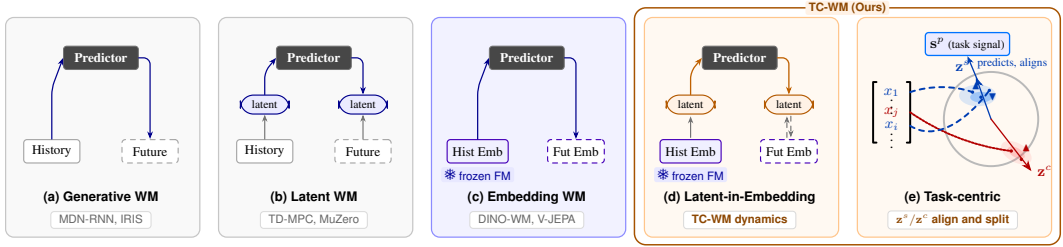
\begin{figure*}[t]
\centering
\resizebox{\textwidth}{!}{%
\begin{tikzpicture}[
    >=Stealth,
    font=\sffamily,
    line cap=round,
    line join=round,
    panel/.style={draw=#1!45, line width=0.75pt, rounded corners=5pt, fill=#1!5},
    module/.style={draw=black!35, line width=0.7pt, rounded corners=2pt, fill=white,
        minimum width=1.20cm, minimum height=0.50cm, align=center, font=\footnotesize},
    dashedmodule/.style={module, dashed, fill=white},
    embed/.style={draw=blue!45!violet, line width=0.7pt, rounded corners=2pt, fill=blue!6,
        minimum width=1.20cm, minimum height=0.50cm, align=center, font=\footnotesize},
    dashedembed/.style={embed, dashed, fill=white},
    latent/.style={draw=blue!55!black, line width=0.7pt, rounded corners=8pt, fill=blue!8,
        minimum width=1.04cm, minimum height=0.36cm, align=center, font=\scriptsize},
    pred/.style={draw=black!55, fill=black!72, text=white, rounded corners=2pt,
        minimum width=1.60cm, minimum height=0.50cm, align=center, font=\footnotesize\bfseries},
    arr/.style={draw=blue!55!black, line width=0.75pt, -{Stealth[length=1.65mm]}},
    softarr/.style={draw=black!55, line width=0.7pt, -{Stealth[length=1.55mm]}},
    taskarr/.style={draw=orange!75!black, line width=0.8pt, -{Stealth[length=1.7mm]}},
    guide/.style={draw=black!20, line width=0.55pt, dashed},
    branch/.style={font=\sffamily\bfseries\footnotesize, align=center},
    ref/.style={draw=black!15, rounded corners=2pt, fill=white, inner xsep=5pt, inner ysep=2pt,
        font=\sffamily\scriptsize, text=black!60, align=center},
    badge/.style={rounded corners=1.5pt, inner xsep=2pt, inner ysep=1pt,
        font=\sffamily\scriptsize, align=center},
]

\begin{scope}[shift={(0,0)}]
    \node[panel=gray, minimum width=4.15cm, minimum height=4.7cm, anchor=south west] at (0,-1.10) {};
    \node[pred] (ap) at (2.05,2.75) {Predictor};
    \node[module] (ah) at (1.00,0.75) {History};
    \node[dashedmodule] (af) at (3.10,0.75) {Future};
    \draw[arr, rounded corners=4pt] (ah.north) -- (1.00,2.10) -- (ap.west);
    \draw[arr, rounded corners=4pt] (ap.east) -- (3.10,2.75) -- (af.north);
    \node[branch] at (2.05,-0.24) {(a) Generative WM};
    \node[ref] at (2.05,-0.72) {MDN-RNN, IRIS};
\end{scope}

\begin{scope}[shift={(4.55,0)}]
    \node[panel=gray, minimum width=4.15cm, minimum height=4.7cm, anchor=south west] at (0,-1.10) {};
    \node[pred] (bp) at (2.05,2.75) {Predictor};
    \node[module] (bh) at (1.00,0.75) {History};
    \node[dashedmodule] (bf) at (3.10,0.75) {Future};
    \node[latent] (bl) at (1.00,1.75) {latent};
    \node[latent] (br) at (3.10,1.75) {latent};
    \draw[softarr] (bh.north) -- (bl.south);
    \draw[arr, rounded corners=4pt] (bl.north) -- (1.00,2.45) -- (bp.west);
    \draw[arr, rounded corners=4pt] (bp.east) -- (3.10,2.75) -- (br.north);
    \draw[softarr] (bf.north) -- (br.south);
    \node[branch] at (2.05,-0.24) {(b) Latent WM};
    \node[ref] at (2.05,-0.72) {TD-MPC, MuZero};
\end{scope}

\begin{scope}[shift={(9.10,0)}]
    \node[panel=blue, minimum width=4.15cm, minimum height=4.7cm, anchor=south west] at (0,-1.10) {};
    \node[pred] (cp) at (2.05,2.75) {Predictor};
    \node[embed] (ch) at (1.00,0.75) {Hist Emb};
    \node[dashedembed] (cf) at (3.10,0.75) {Fut Emb};
    \node[badge, fill=blue!9, text=blue!60!black] at (1.00,0.25) {\Snowflake\ frozen FM};
    \draw[arr, rounded corners=4pt] (ch.north) -- (1.00,2.10) -- (cp.west);
    \draw[arr, rounded corners=4pt] (cp.east) -- (3.10,2.75) -- (cf.north);
    \node[branch] at (2.05,-0.24) {(c) Embedding WM};
    \node[ref] at (2.05,-0.72) {DINO-WM, V-JEPA};
\end{scope}

\begin{scope}[shift={(13.65,0)}]
    \node[panel=orange, minimum width=4.15cm, minimum height=4.7cm, anchor=south west] at (0,-1.10) {};
    \node[pred] (dp) at (2.05,2.75) {Predictor};
    \node[embed] (dh) at (1.00,0.75) {Hist Emb};
    \node[dashedembed] (df) at (3.10,0.75) {Fut Emb};
    \node[badge, fill=blue!9, text=blue!60!black] at (1.00,0.25) {\Snowflake\ frozen FM};
    \node[latent, fill=orange!10, draw=orange!70!black] (dl) at (1.00,1.75) {latent};
    \node[latent, fill=orange!10, draw=orange!70!black] (dr) at (3.10,1.75) {latent};
    \draw[softarr, shorten <=1pt, shorten >=1pt] (dh.north) -- (dl.south);
    \draw[taskarr, rounded corners=4pt] (dl.north) -- (1.00,2.45) -- (dp.west);
    \draw[taskarr, rounded corners=4pt] (dp.east) -- (3.10,2.75) -- (dr.north);
    \draw[softarr, dashed, shorten <=1pt, shorten >=1pt] ([xshift=-1.2pt]df.north) -- ([xshift=-1.2pt]dr.south);
    \draw[softarr, dashed, shorten <=1pt, shorten >=1pt] ([xshift=1.2pt]dr.south) -- ([xshift=1.2pt]df.north);
    \node[branch] at (2.05,-0.24) {(d) Latent-in-Embedding};
    \node[ref, text=orange!55!black] at (2.05,-0.72) {\textbf{TC-WM dynamics}};
\end{scope}

\begin{scope}[shift={(18.20,0)}]
    \node[panel=orange, minimum width=4.15cm, minimum height=4.7cm, anchor=south west] at (0,-1.10) {};
    \begin{scope}[shift={(0.35,0.18)}, scale=0.72]
        \node[draw=blue!55!cyan, thick, fill=blue!8, rounded corners=2pt,
              inner sep=3pt, font=\small]
              (sp) at (2.8,3.98) {$\mathbf{s}^p$ \scriptsize(task signal)};

        \draw[very thick] (0.1,2.85) -- (0,2.85) -- (0,1.05) -- (0.1,1.05);
        \draw[very thick] (1.0,2.85) -- (1.1,2.85) -- (1.1,1.05) -- (1.0,1.05);
        \node[font=\small, text=blue!65!cyan!70!black] at (0.55,2.55) {$x_1$};
        \node[font=\small] at (0.55,2.25) {$\vdots$};
        \node[font=\small, text=red!70!black] at (0.55,1.95) {$x_j$};
        \node[font=\small, text=blue!65!cyan!70!black] at (0.55,1.55) {$x_i$};
        \node[font=\small] at (0.55,1.25) {$\vdots$};

        \def\cx{3.5}
        \def\cy{1.75}
        \def\cr{1.4}
        \draw[very thick, gray!55] (\cx,\cy) circle (\cr);

        \fill[blue!55!cyan, opacity=0.14] (3.1,2.45) ellipse (0.55 and 0.32);
        \fill[blue!55!cyan, opacity=0.28] (3.2,2.4) ellipse (0.32 and 0.16);
        \fill[blue!65!cyan!70!black] (2.95,2.55) circle (0.05);
        \fill[blue!65!cyan!70!black] (3.2,2.4) circle (0.05);
        \fill[blue!65!cyan!70!black] (3.45,2.6) circle (0.05);
        \node[font=\scriptsize, text=blue!65!cyan!70!black] at (3.05,2.75) {$\blacktriangle$};
        \node[font=\scriptsize, text=blue!65!cyan!70!black] at (3.55,2.4) {$\blacktriangledown$};

        \begin{scope}[rotate around={-30:(4.25,0.75)}]
            \fill[red!65, opacity=0.12] (4.25,0.75) ellipse (0.5 and 0.28);
            \fill[red!65, opacity=0.24] (4.25,0.75) ellipse (0.28 and 0.14);
        \end{scope}
        \fill[red!70!black] (4.1,0.8) circle (0.05);
        \fill[red!70!black] (4.4,0.55) circle (0.05);
        \node[font=\scriptsize, text=red!70!black] at (4.55,0.9) {$\blacktriangle$};

        \draw[thick, blue!65!cyan!70!black, ->] (\cx,\cy) -- (3.05,3.0)
            node[above left, font=\small] {$\mathbf{z}^s$};
        \draw[thick, red!70!black, ->] (\cx,\cy) -- (4.7,0.2)
            node[right, font=\small] {$\mathbf{z}^c$};

        \draw[blue!65!cyan!70!black, very thick, dashed]
            (1.1,2.55) .. controls (1.7,2.75) and (2.4,2.75) .. (2.95,2.55);
        \draw[red!70!black, very thick, densely dotted]
            (1.1,1.95) .. controls (1.9,1.4) and (2.9,0.95) .. (4.1,0.8);
        \draw[blue!65!cyan!70!black, very thick, dashed]
            (1.1,1.55) .. controls (1.8,1.1) and (2.7,1.6) .. (3.45,2.6);

        \draw[blue!65!cyan!70!black, thick, -{Stealth[length=1.8mm]}]
            (3.1,2.9) -- (sp.south);
        \node[font=\scriptsize, text=blue!65!cyan!70!black, anchor=west, inner sep=1pt]
            at (3.04,3.40) {predicts, aligns};
    \end{scope}

    \node[branch] at (2.05,-0.24) {(e) Task-centric};
    \node[ref, text=orange!55!black] at (2.05,-0.72)
        {$\mathbf{z}^s/\mathbf{z}^c$ \textbf{align and split}};
\end{scope}

\draw[orange!65!black, line width=0.9pt, rounded corners=7pt]
    (13.45,-1.25) rectangle (22.55,3.82);
\node[badge, fill=orange!10, text=orange!55!black, font=\sffamily\scriptsize\bfseries]
    at (18.00,3.82) {TC-WM (Ours)};

\end{tikzpicture}%
}%

\caption{Comparison of world model paradigms (a)--(c) and our task-centric refinement (d)--(e). Detailed related work can be found in Section~\ref{sec:related-work-extended}.}
\label{fig:four_panel_unified}
\vspace{-0.7em}
\end{figure*}

An agent that can simulate its environment can plan, generalise, and act efficiently from limited experience, a long-standing goal that motivates research on \emph{world models}~\citep{sutton1991dyna,ha2018world,Hafner2020Dream,hansen2024tdmpc}. Recent progress has crystallised the field into three architectural branches that differ in the level at which dynamics are predicted (\Cref{fig:four_panel_unified}). \textbf{\textit{Generative}} models~(a) predict future observations directly in pixel space~\citep{ha2018world}, providing rich visualisation but at heavy computational cost. \textbf{\textit{Latent}} models~(b) instead learn a compact latent representation and predict dynamics within it~\citep{hafner2021mastering,hafner2023mastering,hafner2025mastering,hansen2022temporal,hansen2024tdmpc,schrittwieser2020mastering}, with reward signals (\eg, in TD-MPC and MuZero) often shaping the latent into a task-relevant representation. \textbf{\textit{Embedding}} models~(c), most recently exemplified by DINO-WM and V-JEPA~\citep{assran2025v,zhou2025dinowm,baldassarre2025back}, skip the representation-learning step by predicting directly in the frozen feature space of a pre-trained foundation model, leveraging its broad semantic priors for zero-shot generalisation.

Generally, pretrained visual embeddings are useful not because they form ideal state representations, but because they provide a strong semantic coordinate system from which such states can be extracted. This view is aligned with recent representation-alignment results in generative modelling (\eg, REPA~\citep{yu2025representation}), which shows that tying generative latents to frozen visual encoders simplifies high-dimensional generation. For world models, we therefore build dynamics \emph{on top of} foundation embeddings rather than from pixels alone, inheriting object- and scene-level structure before learning temporal prediction. 

The challenge is that foundation embeddings are optimised for broad semantic coverage rather than for world dynamics or specific planning tasks~\citep{Hafner2020Dream,hu2023toward,zhang2025world}: they encode textures, lighting, and background details that do not affect control. A robot navigating a room needs spatial layout and movable-object dynamics, not wall colour or sofa texture. When generic embeddings serve directly as world-model latents, the model spends capacity on irrelevant variation, hurting modelling efficiency and control interpretability~\citep{schneider2024surprising}. This burden is most acute under \emph{high-dimensional action spaces}: \texttt{Maze} or \texttt{Push-T} use only 2-D actions, but contact-rich \texttt{Robomimic} requires a 7-DoF arm with a $43$-D proprioceptive state (\Cref{tab:dataset-dims}), where every irrelevant latent dimension becomes a possible factor of representation-collapse and wasted planning capacity. The task-centric fix, pursued broadly in representation learning~\citep{lesort2018state,scholkopf2021toward,locatello2020object,ho2022people} and implemented by reward-guided latents in, \eg, TD-MPC2 and MuZero~\citep{hansen2024tdmpc,schrittwieser2020mastering}, is to retain only the information needed to predict task-relevant futures.

We introduce {\method} (\Cref{fig:four_panel_unified}d--e), a task-centric world model designed around this decomposition. Instead of predicting pixels or rolling out the full foundation embedding, {\method} treats the embedding as a reusable visual scaffold and learns compact latent dynamics above it. A designated subspace aligns with the environment's physical state, while the complementary subspace remains anchored to the pretrained embedding via reconstruction, preserving useful visual structure while filtering nuisance variation. This design yields an \textbf{\textit{identifiability}} guarantee of task-relevant representation of world dynamics, validated by our linear probing and ablations. 

Once the task-centric representation is ready, it supports \emph{a wide suite of} downstream policy-learning control, including test-time planning with model predictive control~\citep{draeger1995model} for goal-reaching tasks, a latent diffusion planner~\citep{xie2025latent} for manipulation tasks, and model-free reinforcement learning~\citep{haarnoja2018soft} for locomotion tasks. 

\begin{figure}[!h]
    \centering
    \includegraphics[width=\linewidth]{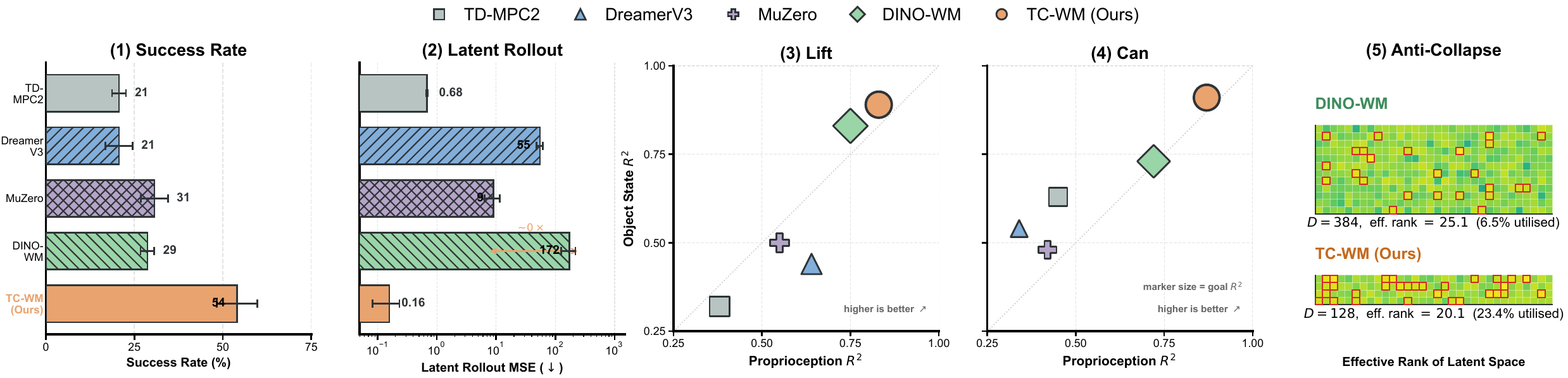}
    \caption{Experiment results on \texttt{Robomimic}. \textbf{(1)} Success rate. \textbf{(2)} Latent-rollout MSE. \textbf{(3)--(4)} Linear probes on \texttt{Lift} and \texttt{Can}. \textbf{(5)} Anti-collapse: {\method} has better utilisation of latent space.}
    \label{fig:teaser}
\end{figure}

Empirically, {\method} improves both world model quality, reward-free offline planning and control across 9 benchmarks; \Cref{fig:teaser} highlights the \texttt{Robomimic} setting with a high-dimensional action space, where the gains are largest, the learned latent linearly recovers physical states, and our design prevents representation collapse compared with models directly building on visual embeddings.

\section{Method}
\begin{figure}[!t]
    \centering
    \includegraphics[width=\linewidth]{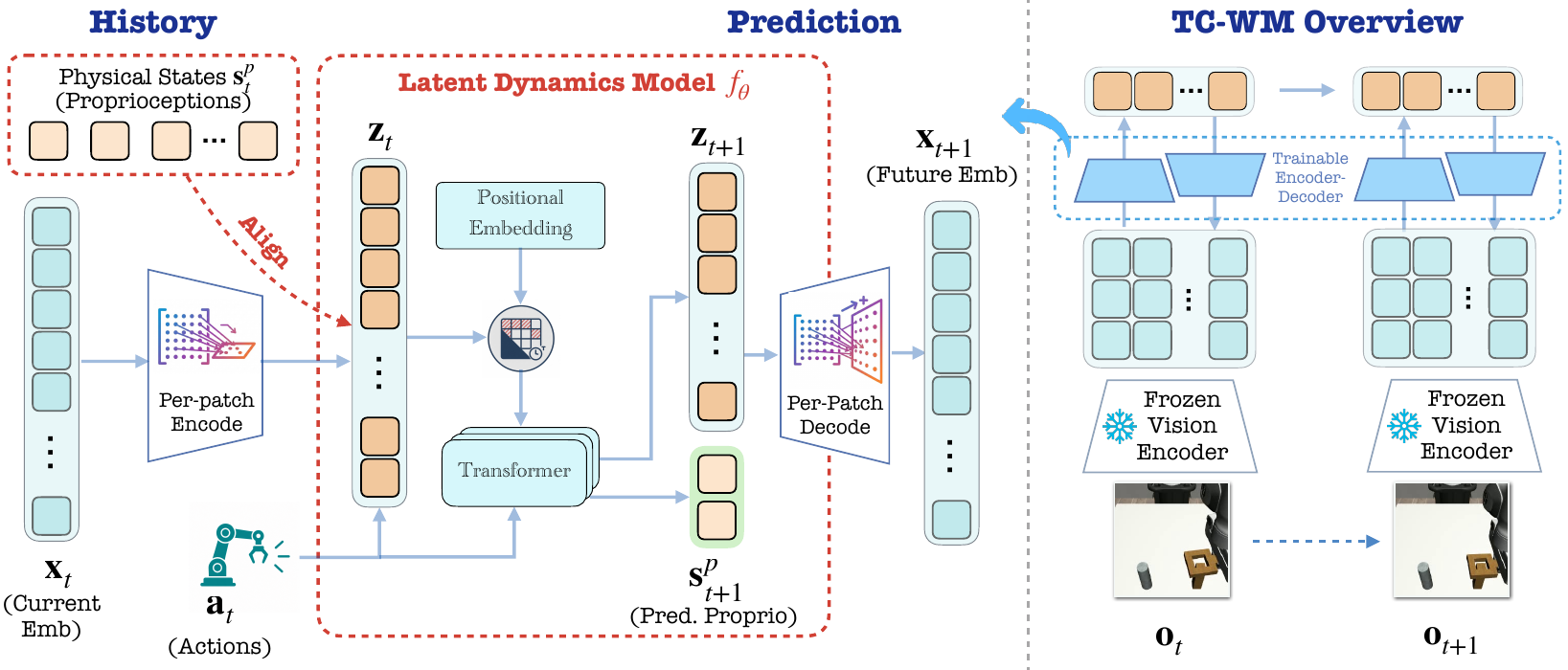}
    \caption{We present \textbf{{\method}}, a method for training a world model by extracting a compact, task-centric latent dynamics from pretrained visual embeddings of image frames using the task-centric side information, \eg, proprioception.}
    \label{fig:architecture}
\end{figure}

\subsection{Problem Setup}
We formulate the problem as a partially observable Markov decision process (POMDP)~\citep{kaelbling1998planning} defined by $(\mathcal{S}, \mathcal{O}, \mathcal{A}, p_{\text{env}})$, where $\mathcal{S}$ denotes the latent environment state space, $\mathcal{O}$ the observation space, and $\mathcal{A}$ a continuous action space. The environment evolves according to latent dynamics $p_{\text{env}}(\ss_{t+1} \mid \ss_t, \aa_t)$, while observations $\oo_t \in \mathcal{O}$ are generated from latent states $\ss_t$ through an unknown observation process. We assume access to a low-dimensional \emph{proprioceptive} signal $\ss^p_t \in \mathbb{R}^d$, which captures the agent's physical configuration (\eg, joint positions, end-effector pose, gripper state). This signal represents a partial, directly observable projection of the full state $\ss_t$.

Given an offline dataset of visual trajectories $\mathcal{D}=\{(\oo_t,\aa_t, \ss^p_t)\}_{t=0}^T$, where visual observations can be further encoded into embeddings $\mathbf{x}_t$, our goal is to learn a world model that infers a latent representation $\zz_t$ from observation histories and models its dynamics. We aim for the learned latent sequence $[\zz_0,\ldots,\zz_T]$ to be \emph{task-centric}, meaning that it preserves information necessary to predict future task-relevant dynamics under actions, which is formally defined below.
\begin{definition}[Task-Centric Representation] \label{def:task-centric}
A latent representation $\zz_t = f(\oo_{\le t}, \aa_{<t})$ is \emph{task-centric} with respect to proprioceptive state $\ss^p_t$ if it satisfies the following two conditions.
\begin{equation} \notag
\resizebox{\linewidth}{!}{$
\underbrace{
p_{\text{env}}(\ss^p_{t+H} \mid \oo_{\le t}, \aa_{<t}, \aa_{t:t+H-1})
\,=\, p_{\text{env}}(\ss^p_{t+H} \mid \zz_t, \aa_{t:t+H-1})
}_{\substack{\text{\footnotesize \textbf{Sufficiency:} for any action sequence}\\[-0.2ex]\text{\footnotesize replacing history by } \zz_t \text{\footnotesize\ leaves the future unchanged}}}
\qquad
\underbrace{
p_{\text{env}}(\ss^p_t \mid \zz_t)
\,=\, p_{\text{env}}(\ss^p_t \mid \oo_{\le t}, \aa_{<t})
}_{\substack{\text{\footnotesize \textbf{Task-Aligned:}}\\[-0.2ex]\text{\footnotesize preserve the current proprioceptive state}}}
$}
\end{equation}
\end{definition}

\subsection{Task-Centric World Model ({\method})} \label{sec:overall-model}
\paragraph{Overview.} {\method} learns a compact rollout state $\zz_t$ on top of frozen visual embeddings, using proprioception to identify the dimensions needed for physical dynamics. Figure~\ref{fig:architecture} illustrates the pipeline, whose main components are:
\begin{equation}
\begingroup
\renewcommand{\arraystretch}{1.3}
\begin{array}{@{}l@{\quad}l@{\;}c@{\;}l@{\qquad}l@{}}
\text{Joint embedding} &
\xx_t &=&
\big[f^{o}_{\text{emb}}(\oo_t),\, f^{p}_{\text{emb}}(\ss^p_t)\big]
& \triangleright\ \text{\textcolor{gray}{\textit{Frozen visual + trainable proprio}}} \\

\text{Latent encoder} &
\zz_t &\sim&
q(\zz_t \mid \xx_t)
& \triangleright\ \text{\textcolor{gray}{\textit{Linear projection to compact latent}}} \\

\text{TC alignment} &
\zz^s_t &\leftrightarrow&
\ss^p_t,\ \zz_t{=}[\zz^s_t,\zz^c_t]
& \triangleright\ \text{\textcolor{gray}{\textit{Align $\zz^s$ with proprio}}} \\

\text{Latent dynamics} &
\hat{\zz}_{t+1} &=&
f_{\theta}\big(\zz_{t\text{-}H:t},\, \aa_{t\text{-}H:t}\big)
& \triangleright\ \text{\textcolor{gray}{\textit{Predicts next latent state}}} \\

\text{TC dynamics} &
\hat{\ss}^{p}_{t+1} &=&
f^{p}_{\eta}\big(\zz_{t\text{-}H:t},\, \aa_{t\text{-}H:t}\big)
& \triangleright\ \text{\textcolor{gray}{\textit{Predicts next proprioception}}} \\

\text{Embedding decoder} &
\hat{\xx}_t &=&
f_{\text{dec}}(\zz_t)
& \triangleright\ \text{\textcolor{gray}{\textit{Reconstructs visual embedding}}}
\end{array}
\endgroup
\label{eq:overall-model}
\end{equation}
Here $H$ is the prediction horizon, $\oo_t$ the image observation, and $\ss^p_t$ the observed proprioceptive signal. The embedding $\xx_t$ serves as the visual anchor, while $\zz_t$ is the latent state optimised for prediction and planning. We introduce each module below.

\paragraph{$\triangleright$ Joint Embedding.}
At each time step, a frozen encoder $f^o_{\text{emb}}$, such as DINOv2~\citep{oquab2023dinov2}, maps the image $\oo_t$ to a visual feature, while a trainable linear layer $f^p_{\text{emb}}$ maps proprioception $\ss^p_t$. Their concatenation $\xx_t=[f^o_{\text{emb}}(\oo_t), f^p_{\text{emb}}(\ss^p_t)]$ is not the rollout state. It provides a structured input to the latent encoder: frozen visual features contribute spatially coherent, object-centric cues~\citep{zhou2025dinowm,baldassarre2025back,yu2025representation}, and proprioception supplies the physical coordinate injection, which serves as side information.

\paragraph{$\triangleright$ Latent Encoder.}
To further distill the visual embeddings, {\method} treats them as a scaffold rather than directly modeling temporal dynamics within them. Specifically, we introduce an encoder $q(\zz_t \mid \xx_t)$ that transforms the joint embedding $\xx_t$ into a latent representation $\zz_t$, which is further trained to satisfy Definition~\ref{def:task-centric}. This projection is the central abstraction step: it retains the semantic directions useful for reconstructing and predicting the scene, but compresses away dimensions that are not useful for future physical state prediction.

\paragraph{$\triangleright$ TC Alignment.}
Directly aligning the full latent space with $\ss^p$ would discard the rich semantic content captured by the visual embedding. We therefore perform a \emph{partial} alignment: we factorise $\zz_t = [\zz_t^{s}, \zz_t^{c}]$ and align only the state-relevant block $\zz_t^{s}$ with the proprioceptive state $\ss^p_t$ via an InfoNCE-style contrastive objective~\citep{oord2018representation}, leaving the complementary block $\zz_t^{c}$ free to absorb residual semantic structure (object appearance, background, etc.) so that the embedding-reconstruction loss introduced below can still be satisfied. Critically, this contrastive term plays a stronger role than mere prediction: prior conditional generative identifiability frameworks~\citep{khemakhem2020variational,zimmermann2021contrastive} \emph{passively} \textit{\textbf{assume}} \textit{mechanism diversity}, that is, distinct physical states inducing distinct conditionals, as a property of the data-generating process. \method{}'s alignment instead \emph{actively} \textbf{produces} this diversity on $\zz_t^{s}$ by pushing latents tied to different proprioceptive states apart, turning what was a passive assumption (cf.~A2 of \Cref{thm:blk idn}) into a learned property of the encoder.
\begin{equation}\label{eq:infonce}
    \mathcal{L}_{\text{align}} 
= - \log 
\frac{
    \exp(\text{sim}(g_\phi(\zz_t^s), h_\psi(\ss^p_t))/\tau)
}{
    \sum_{\ss^-_t}
    \exp(\text{sim}(g_\phi(\zz_t^s), h_\psi(\ss^{p-}_t))/\tau)
},
\end{equation}
where $g_\phi$, $h_\psi$ are small projection heads, $\text{sim}(\cdot,\cdot)$ is cosine similarity, $\tau$ is the temperature, positive pairs $(\zz_t^s, \ss^p_t)$ are drawn from the same timestep, and negatives $\ss^{p-}_t$ from other samples in the batch. In practice, $g_\phi$ is implemented as a \emph{linear} map $\mathbb{R}^{d_z}\!\to\!\mathbb{R}^{d_s}$ with an $\ell_1$ penalty on its weight matrix, so $\zz_t^{s}$ is automatically \emph{selected} by sparsity rather than pre-sliced from $\zz_t$; $d_s$ functions only as an upper bound on the width of the task-centric block.

\paragraph{$\triangleright$ Latent Dynamics.}
We adopt the ViT~\citep{dosovitskiy2020image} as the transition architecture $f_\theta$ to predict $\hat{\zz}_{t+1}=f_\theta(\zz_{t-H:t},\aa_{t-H:t})$, which models the temporal dependencies of states and actions by minimizing a predictive consistency loss
\begin{equation} \nonumber
    \mathcal{L}^z_{\text{dyn}} = \|\hat{\zz}_{t+1} - q(\zz_{t+1}\mid\xx_{t+1})\|_2^2
\end{equation} 
which encourages the transition to capture dynamics in the projected latent space rather than at the raw embedding level.

\paragraph{$\triangleright$ TC Dynamics.}
Beyond instantaneous alignment, we predict these proprioceptive variables directly from the history to encourage the latent space to capture task-centric dynamics:
\begin{equation} \nonumber
    \mathcal{L}^s_{\text{dyn}} = \|f^p_\eta(\zz_{t-H:t},\aa_{t-H:t}) - \ss^p_{t+1}\|_2^2.
\end{equation} 
We predict raw (but standardised) proprioception rather than its embedding, which avoids trivial solutions where the state-embedding layer $f^p_{\text{emb}}$ collapses to match its own output.

\paragraph{$\triangleright$ Embedding Decoder.}
A trainable linear decoder $f_{\text{dec}}$ reconstructs the visual embedding $\xx_t$ from $\zz_t$. This term acts as a \emph{regulariser} that anchors the compact latent to the foundation-embedding manifold, keeping the latent space smooth, well-spread across samples, and free of the degenerate solutions that compact latents are otherwise prone to:
\begin{equation}
\hat{\xx}_t = f_{\text{dec}}(\zz_t), \qquad
\mathcal{L}_{\text{rec}} = \left\| \hat{\xx}_t - \xx_t \right\|_2^2 .
\label{eq:embedding-decoder}
\end{equation}
In our view, the embedding of a visual foundation model, \eg, DINO~\citep{oquab2023dinov2} or V-JEPA~\citep{assran2023self}, can be a better reconstruction target than raw pixels (which generative world models typically use~\citep{Hafner2020Dream,hafner2023mastering,alonso2024diffusion}): it offers a denoised supervision signal that promotes a stable, compact latent, as verified in Section~\ref{sec:ana_lin}.

\paragraph{Visual Decoder.}
For visualization \emph{only}, an auxiliary decoder maps predicted embeddings to images, $\hat\oo_t=f_{\text{vis}}(\hat\xx_t)$, with $\mathcal{L}_{\text{img}}=\|\hat\oo_t-\oo_t\|_2^2$ separated from world-model optimization.

The overall training objective combines dynamics prediction and representation alignment:
\[
\mathcal{L}_{\text{all}} = 
\mathcal{L}^z_{\text{dyn}} + \mathcal{L}^s_{\text{dyn}} +
\lambda_{\text{align}} \mathcal{L}_{\text{align}} + \lambda_{\text{rec}} \mathcal{L}_{\text{rec}},
\]
where $\lambda_{\text{align}}$ balances alignment and dynamics, and $\lambda_{\text{rec}}$ governs the embedding reconstructions.

\begin{figure}[!t]
\centering
\includegraphics[width=\linewidth]{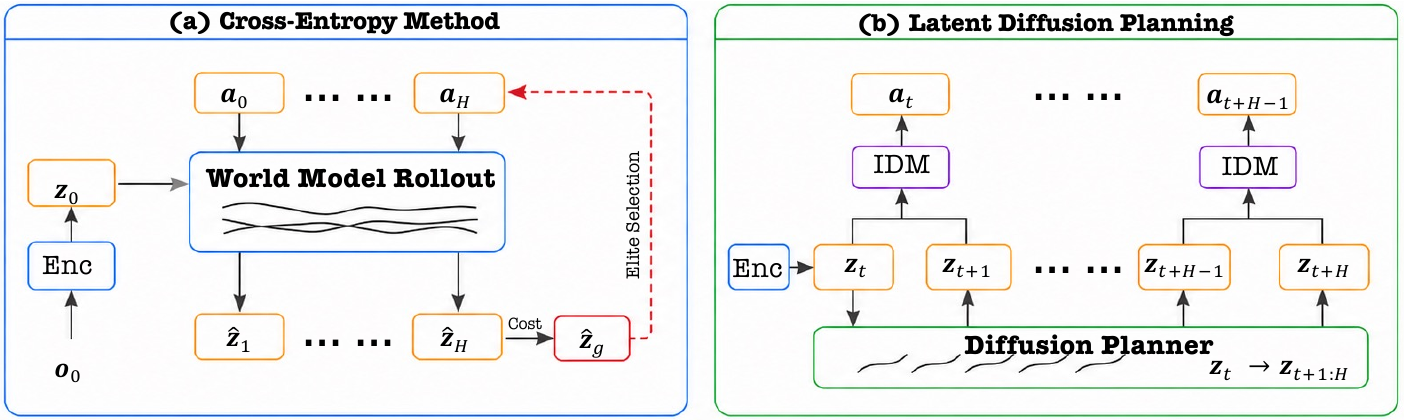}
\caption{Planning algorithms used in {\method}. \textbf{(a)~Cross-Entropy Method (CEM)} for navigation: the world model rolls out $K$ sampled action sequences, scores each rollout against a goal latent $\zz_g$ via cost $\mathcal{C}$, and refines the action distribution by elite selection. \textbf{(b)~LDP planning} for manipulation: a diffusion planner generates future latent trajectories $[\zz_{t{+}1},\ldots,\zz_{t{+}H}]$, and an inverse dynamics model (IDM) recovers executable actions from consecutive latent pairs.}
\label{fig:ldp-pipeline}
\end{figure}

\subsection{Test-Time Planning and Control with {\method}}
\label{sec:plan-intro}
To validate whether the task-centric representations can be translated into better downstream planning, we evaluate the learned world model on visual goal-reaching tasks performed entirely in the latent space $\mathcal{Z}$. 
Given an initial observation $\oo_0$ and a goal observation $\oo_g$, we encode them as $\zz_0 = q(\zz_0 \mid \xx_0)$ and $\zz_g = q(\zz_g \mid \xx_g)$. 
Planning is then formulated as minimizing a latent-space objective that measures the distance between the predicted latent state and the goal state:
\[
\mathcal{L}_{\text{plan}}(\aa_{0:T}) 
= 
\| f_\theta^H(\zz_0, \aa_{0:T}) - \zz_g \|_2^2.
\]
For tasks with low-dimensional action spaces, such as \textit{navigation}, we adopt model predictive control (MPC)~\citep{draeger1995model} with the cross-entropy method (CEM) to iteratively refine action sequences based on predicted rollouts by $\mathcal{L}_{\text{plan}}(\aa_{0:T})$ (\Cref{fig:ldp-pipeline}a). For high-dimensional, contact-rich \textit{manipulation}, we use latent diffusion planning (LDP)~\citep{xie2025latent} over {\method}'s task-centric latents (\Cref{fig:ldp-pipeline}b). LDP trains a diffusion planner for future latent trajectories and an inverse dynamics model (IDM) for executable actions:
\begin{equation*}
\resizebox{\linewidth}{!}{$\displaystyle
\mathcal{L}_{\text{LDP}}(\psi)=\mathbb{E}_{\tau,\epsilon}\!\left[\|\epsilon_{\psi}(\hat{\zz}_{t+1:t+H};\zz_t,\tau)-\epsilon\|_2^2\right],
\quad
\mathcal{L}_{\text{IDM}}(\xi)=\mathbb{E}_{i,\tau,\epsilon}\!\left[\|\epsilon_{\xi}(\hat{\aa}_{t+i};\hat{\zz}_{t+i},\hat{\zz}_{t+i+1},\tau)-\epsilon\|_2^2\right].
$}
\end{equation*}
At inference, the planner samples $\hat{\zz}_{t+1:t+H}$ by denoising conditioned on $\zz_t$, and the IDM decodes actions from consecutive latent pairs. Compared with the original VAE-based LDP, replacing the encoder with {\method}'s latent space gives a more structured representation for planning; details are in Appendix~\ref{sec:plan_alg}. For continuous control, we couple {\method} with an off-policy actor--critic. The resulting latent $\zz_t$ obtained by {\method} is treated as the observation of a Soft Actor--Critic (SAC) agent~\citep{haarnoja2018soft}, which is then trained from environment interaction with the same protocol used by the baselines. We report the mean episode return across 3 seeds.                

\noindent\textbf{Theoretical Rationale.}
The following theorem formalizes why a compact latent learned on frozen embeddings, once aligned with physical signals, can recover the task-relevant state needed for prediction and planning. We write $L_{\mathbf{v}\mid\mathbf{u}}$ for the conditional linear operator induced by $p(\mathbf{v}\mid\mathbf{u})$, where $[L_{\mathbf{v}\mid\mathbf{u}}\phi](\mathbf{v})=\int p(\mathbf{v}\mid\mathbf{u})\phi(\mathbf{u})\,d\mathbf{u}$. Injectivity of $L$ means that distinct latent states induce distinguishable futures, i.e., the observed dynamics vary across latent states.
\refstepcounter{theorem}\label{thm:tcwm-main}
\begin{tcwmtheorembox}{Theorem~\thetheorem\ \textbf{\textit{(Task-centric identifiability from visual embeddings)}}}
\textit{Suppose images $\oo_t$, frozen embeddings $\xx_t$, and latent variables $\zz_t=[\zz^s_t,\zz^c_t]$ follow the process in~\Cref{sec:overall-model}, and the learned latent world model matches the true joint distribution of adjacent embeddings $\{\xx_{t-1},\xx_t,\xx_{t+1}\}$. Assume:}
\begin{description}[leftmargin=0pt, labelwidth=2.7em, labelsep=0.55em, style=sameline, itemsep=0.2em, topsep=0.2em]
    \item[\textbf{A1.}] \textbf{\textit{(\underline{Contextual dynamics}):}} \textit{$L_{\xx_{t+1:t+H}\mid \zz_t}$ and $L_{\xx_{t-H:t-1}\mid \xx_{t+1:t+H}}$ are injective and bounded.}
    \item[\textbf{A2.}] \textbf{\textit{(\underline{Latent--observed variability}):}} \textit{For any $\zz_t^{(1)}\neq \zz_t^{(2)}$, $p(\xx_t\mid \zz_t^{(1)})\neq p(\xx_t\mid \zz_t^{(2)})$.}
    \item[\textbf{A3.}] \textbf{\textit{(\underline{Differentiability}):}} \textit{There exists a differentiable $F$ such that $F[p_{\xx_t\mid\zz_t}(\cdot\mid\zz_t)]=h_z(\zz_t)$.}
    \item[\textbf{A4.}] \textbf{\textit{(\underline{Partial alignment}):}} \textit{The alignment objective in~\Cref{eq:infonce} is sufficiently minimized.}
\end{description}
\textit{Then the learned $\zz_t$ is block-wise identifiable, and the task-centric $\zz^s_t$ is affine-identifiable:}
\begin{equation*}
\resizebox{\linewidth}{!}{$
\underbrace{
\hzz_t = h_z(\zz_t),\quad h_z \text{ invertible and } C^1
}_{\substack{\text{\footnotesize \textbf{Latent Space Recovery:} the compact latent preserves}\\[-0.15ex]\text{\footnotesize the true world state up to reparameterization}}}
\qquad
\underbrace{
\zz^s_t = \mathbf{A}\hzz^s_t + \mathbf{b}
}_{\substack{\text{\footnotesize \textbf{Task-Centric Recovery:} the aligned block recovers}\\[-0.15ex]\text{\footnotesize physical factors up to an affine map}}}
$}
\end{equation*}
\end{tcwmtheorembox}
\noindent\textit{Proof sketch and intuition.}
The observable joint distribution over neighboring embeddings determines an operator family that is a similarity transform of the diagonal operator indexed by $p(\xx_t\mid \zz_t)$. Injective context operators make this diagonalization unique up to a smooth reparameterization, and the alignment loss selects the task-centric block by tying it to physical variables. Thus, the linear projection and proprioceptive alignment in {\method} can recover a compact, physically meaningful latent (see Appendix~\ref{sec:thm}) rather than only producing successful predictions on the training set.
\section{Experiments} \label{app:exp}
We evaluate {\method} across a diverse set of offline control and visual planning benchmarks to assess whether the learned representations are \emph{task-centric}, \emph{predictively sufficient}, and effective for downstream decision-making. Our evaluation spans navigation, locomotion, and manipulation tasks across simulated and real-world settings. We focus on three complementary aspects: \texttt{(i)} the accuracy and stability of the learned world model, \texttt{(ii)} planning performance under both low- and high-dimensional action spaces, and \texttt{(iii)} the interpretability and structure of the learned latent representations through targeted ablations and analysis.
\subsection{Environments and Tasks}
We evaluate {\method} on 9 offline visual-control environments spanning \emph{navigation}, \emph{locomotion}, and \emph{manipulation} in different metrics.

\paragraph{Datasets.}
\begin{wrapfigure}[14]{r}{0.48\linewidth}
\vspace{-2em}
    \centering
    \includegraphics[width=\linewidth]{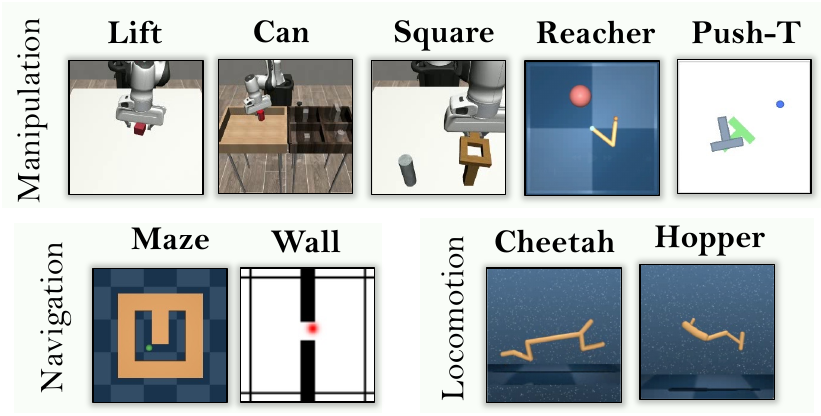}
    \caption{\textbf{Environments used in Experiments}.}
    \label{fig:datasets}
\end{wrapfigure}
As shown in Figure~\ref{fig:datasets}, the benchmark suite includes \texttt{Maze} and \texttt{Wall} for navigation, \texttt{Push-T} and \texttt{RoboMimic}~\cite{mandlekar2021matters} (\texttt{Lift}, \texttt{Can}, \texttt{Square}) for manipulation, and \texttt{Reacher}, \texttt{Cheetah}, and \texttt{Hopper} from DeepMind Control Suite~\citep{tassa2018deepmind} for continuous control. Dataset construction, action/state dimensions, and preprocessing details are provided in Appendix~\ref{sec:data-details}. 
\WFclear
\vspace{1.8em}
\paragraph{Evaluation.}
We evaluate prediction quality using image-space reconstruction, latent rollout error, and SSIM. For goal-reaching tasks, we report success rate (SR); for DMC locomotion tasks, we report episode return.
\paragraph{Choices of Visual Foundation Models.}
We use DINOv2~\citep{oquab2023dinov2} as the default frozen visual encoder, matching DINO-WM for fair comparison. Figure~\ref{fig:arch-ablation} additionally evaluates DINOv3~\citep{simeoni2025dinov3} and Cosmos~\citep{agarwal2025cosmos} as alternative visual foundation models; implementation details are in Appendix~\ref{app:vis_fm}.

\paragraph{Baselines.} We compare against four representative world-model baselines. \textbf{TD-MPC2}~\citep{hansen2024tdmpc} learns compact latent dynamics for model-predictive control, adapted here to replay-only offline training. \textbf{DreamerV3}~\citep{hafner2025mastering} learns a recurrent latent generative world model with image reconstruction and imagination. \textbf{MuZero}~\citep{schrittwieser2020mastering}, a representative task-oriented latent-dynamics method, is reproduced inside our framework as a representation/dynamics trained on the same offline trajectories (\Cref{app:muzero-offline}). \textbf{DINO-WM}~\citep{zhou2025dinowm} predicts future frozen DINO embeddings directly and is the closest embedding-space baseline. All methods use the same offline trajectories of images, actions, and proprioception.

\subsection{Results and Analysis}

\begin{figure}[t]
    \centering
    \includegraphics[width=\linewidth]{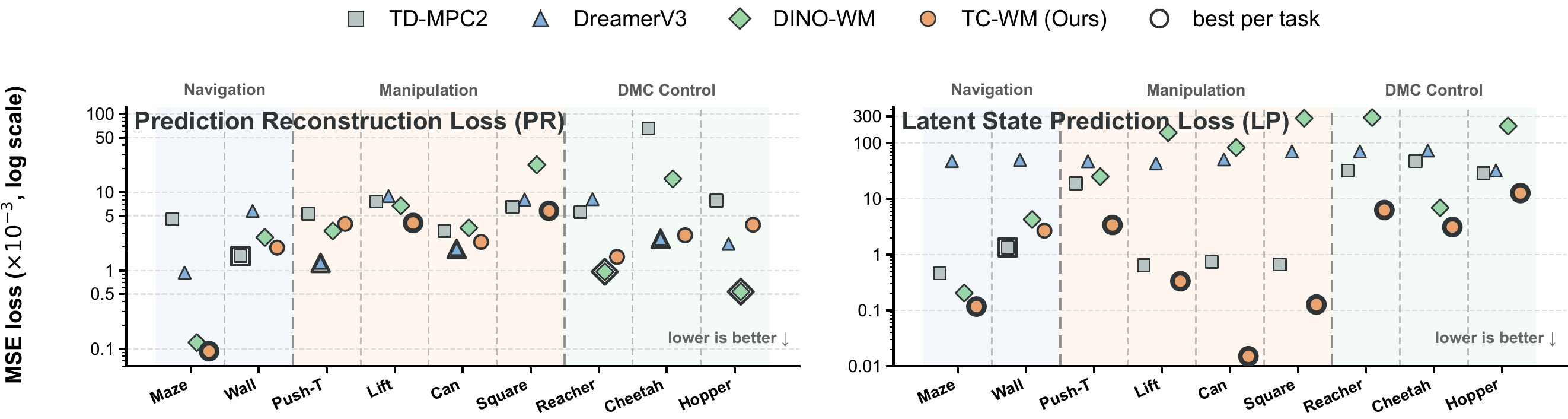}
    \caption{World model prediction accuracy. We report Prediction Reconstruction Loss (PR) and Latent State Prediction Loss (LP) (lower is better) on offline datasets over 300k training steps.}
    \label{fig:pred_loss}
    \vspace{-1em}
\end{figure}

\paragraph{World Model Prediction.}
Figure~\ref{fig:pred_loss} separates image reconstruction from latent rollout accuracy. Image-space error is task-dependent: {\method} is competitive on \texttt{Lift}, \texttt{Square}, and \texttt{Maze}, while baselines perform better on visually complex tasks like \texttt{Push-T}, \texttt{Cheetah}, and \texttt{Hopper}. In contrast, {\method} achieves the lowest latent prediction error on nearly all tasks, especially in manipulation settings where accurate dynamics are crucial. This suggests that compressing frozen visual embeddings preserves key structure while yielding a more stable state space.

\begin{figure}[H]
    \centering
    \includegraphics[width=0.8\linewidth]{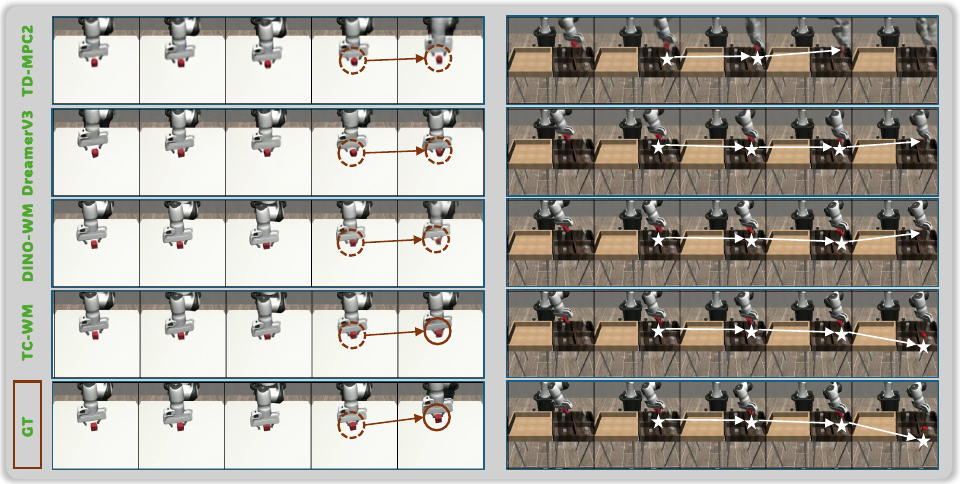}
    \caption{Open-loop rollouts on \texttt{RoboMimic}: \texttt{Lift} (left half) and \texttt{Can} (right half).}
    \label{fig:openloop_robomimic}
\end{figure}

\begin{figure}[H]
    \centering
    \includegraphics[width=\linewidth]{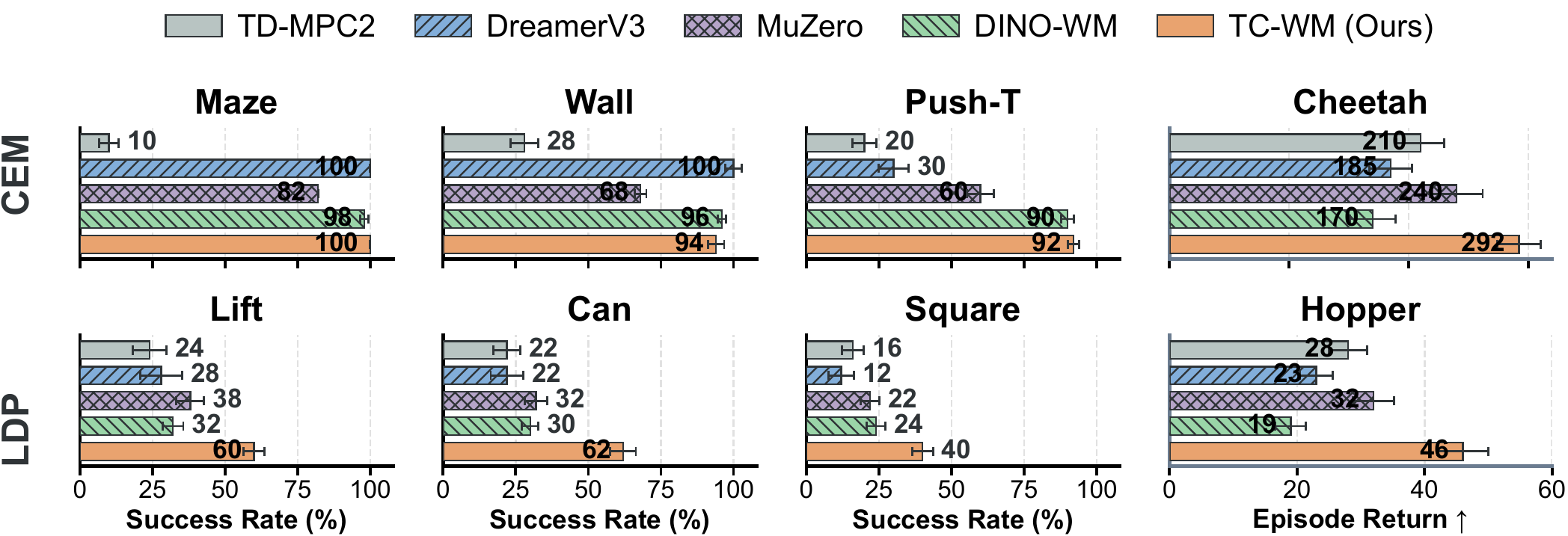}
    \caption{Planning performance of {\method} against TD-MPC2, DreamerV3, MuZero, and DINO-WM. \textit{(Top)} CEM planning on \texttt{Maze}, \texttt{Wall}, \texttt{Push-T} (success rate, \%) and on \texttt{Cheetah} and \texttt{Hopper} (episode return; the rightmost column has a blue-grey background to flag the change of metric). \textit{(Bottom)} LDP planning on \texttt{Lift}, \texttt{Can}, and \texttt{Square}, where high-dimensional action spaces make task-centric latent dynamics especially important. {\method} matches the strongest baseline on the saturated CEM tasks and is the only method that surpasses DINO-WM on every LDP task.}
    \label{fig:exp-plan}
\end{figure}

\begin{figure}[h]
\centering
\includegraphics[width=\linewidth]{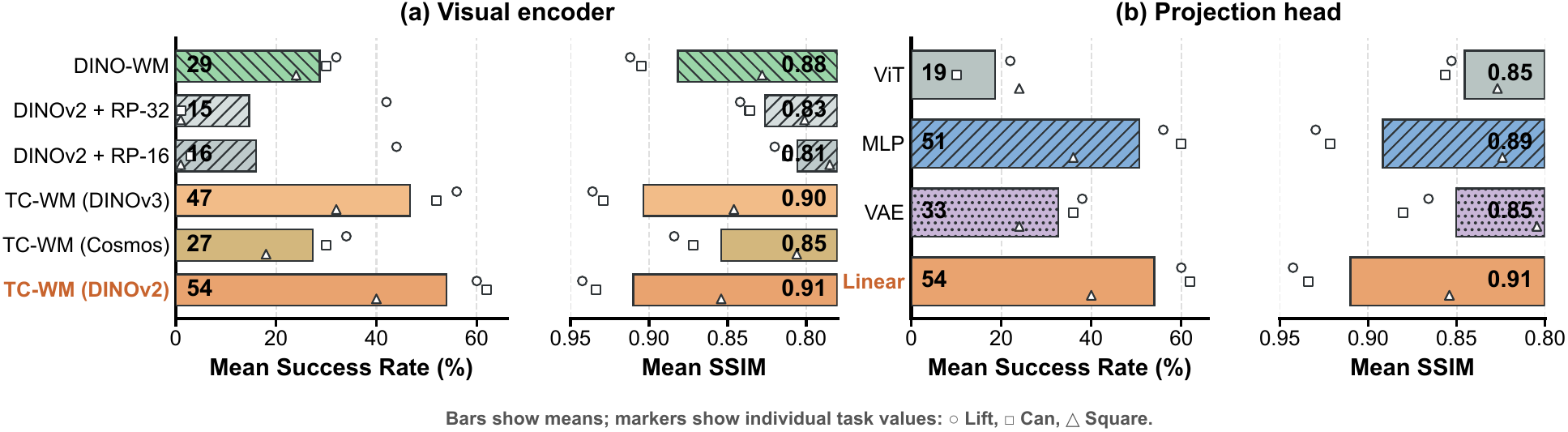}
\vspace{-0.6em}
\caption{Architecture/module ablations on \texttt{RoboMimic}. \textit{(a)}~Visual encoder or latent-source choice and \textit{(b)}~projection-head ablation, each showing success rate on the left and mirrored SSIM on the right. RP denotes randomly projected DINOv2 embeddings. Bars show means over \texttt{Lift}, \texttt{Can}, and \texttt{Square}; markers show individual task values.}
\label{fig:arch-ablation}
\vspace{-1em}
\end{figure}

Open-loop rollouts in Figure~\ref{fig:openloop_robomimic} provide the qualitative counterpart: {\method} better preserves contact events, gripper motion, and object placement over long horizons. The main tradeoff is image-space fidelity on visually complex dynamics, where compact projection can discard low-level details.
\vspace{-2em}
\paragraph{Planning Performance.} \label{app:planning_main_performance}
Figure~\ref{fig:exp-plan} evaluates planning with CEM for low-dimensional control and LDP~\citep{xie2025latent} for high-dimensional manipulation. {\method} is competitive across CEM tasks and obtains the clearest gains under LDP on \texttt{Lift}, \texttt{Can}, and \texttt{Square}, where planning depends on precise long-horizon object interaction. This supports the main empirical claim: task-centric latents matter most when the action space is complex and control-relevant physical state must remain stable.
\vspace{-1em}
\vspace{-1em}
\paragraph{Architecture and Module Ablations.}
Figure~\ref{fig:arch-ablation} summarizes architecture-level ablations on \texttt{RoboMimic}. Random projection performs poorly, showing that dimensionality reduction alone is not enough. DINOv2 and DINOv3 provide the strongest foundation features, while Cosmos is weaker in this setting. Among projection heads, the linear map gives the best balance between planning success and embedding reconstruction.

\begin{figure*}[t]
    \centering
    \includegraphics[width=\linewidth]{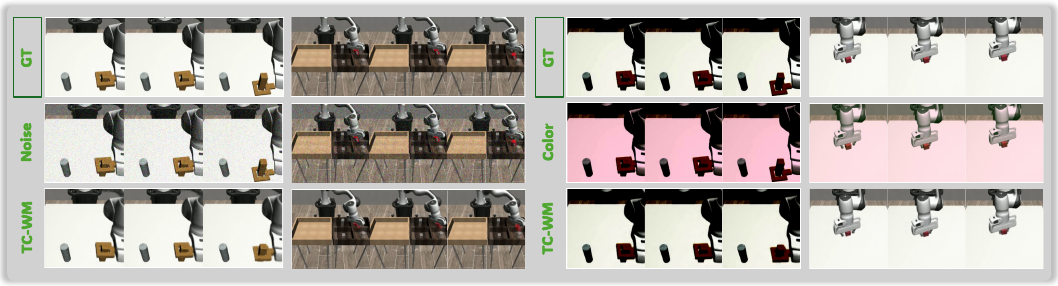}
    \caption{Rollouts under unseen visual perturbations. \emph{Left:} Gaussian noise on \texttt{Square} and \texttt{Can}. \emph{Right:} color jitter on \texttt{Can} and \texttt{Lift}.}
    \label{fig:unseen_perturb}
\end{figure*}

\paragraph{Robustness to unseen visual perturbations.}
Following the test-time perturbation protocol of DINO-WM~\citep{zhou2025dinowm}, we feed each frozen world model inputs corrupted by \emph{Gaussian noise} ($\sigma{=}0.1$) or \emph{color jitter} (per-channel scale $\in[0.8,1.2]$, shift $\sigma{=}0.05$). \Cref{fig:unseen_perturb} shows {\method} recovers the manipulator and object pose under both perturbations, and its relative SSIM drop is consistently smaller than DINO-WM's, \eg, \ $-2.5\%$ vs.\ $-4.8\%$ under noise on \texttt{Lift}.

\subsection{Evaluation on Learned  Representations} \label{sec:ana_lin}
\paragraph{Linear Probing on Latent Representation.}
To validate that {\method} learns task-centric representations, we probe the learned latent space using ground-truth \textit{full} states, including proprioceptive signals, object states, and goal-related indicators. We use the same protocol for all methods: train a linear regressor from each method's latent to the target state variables and report $R^2$ scores. The linear-probing panels in Figure~\ref{fig:teaser} show that {\method} more accurately recovers these task-centric variables, indicating that its latent space is not merely predictive but physically grounded.

\paragraph{Projection Design Rationale.}
Figure~\ref{fig:arch-ablation} also explains why we use a linear projection. Frozen visual embeddings already contain structured semantic directions, so the projection mainly needs to select and align a compact task-centric subspace rather than relearn visual abstraction. More complex projectors can distort this geometry or introduce optimization and bottleneck effects. This empirical trend is consistent with Theorem~\ref{thm:tcwm-main}, which shows that a linear transformation is sufficient to recover the task-centric subspace up to an affine mapping.

\begin{figure}[h]
    \centering
    \includegraphics[width=\linewidth]{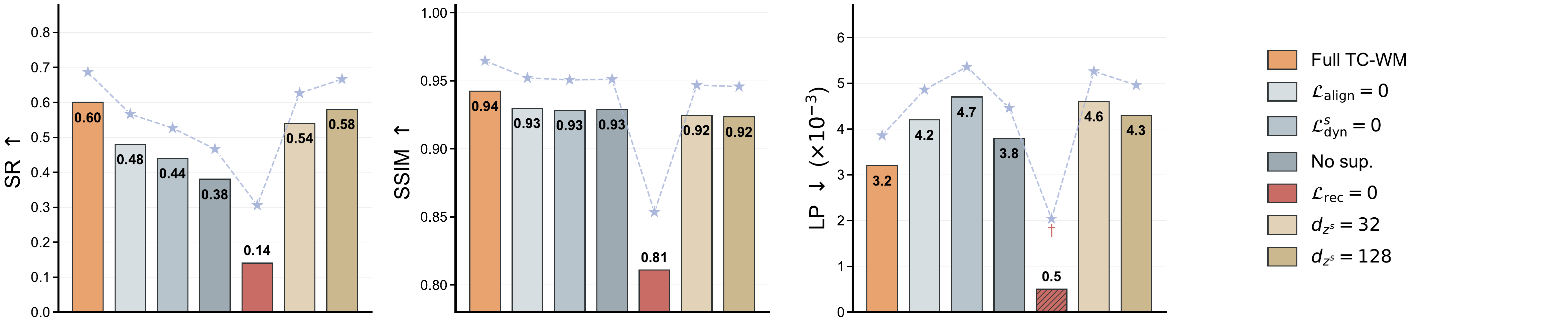}
    \caption{Ablation on loss components and latent split dimension (\texttt{Lift}, 100 epochs). $\mathcal{L}_{\text{rec}}{=}0$ collapses both planning (SR) and visual fidelity (SSIM); proprioceptive supervision ($\mathcal{L}_{\text{align}}$, $\mathcal{L}^s_{\text{dyn}}$) is essential for planning; $\zz^s$ split dimension has modest effect. \protect{$\dagger$} Without embedding reconstruction, the latent collapses and trivially predicts itself, so LP is artificially low.}
    \label{fig:loss-ablation}
\end{figure}

\paragraph{Ablation on Loss Components.}
Figure~\ref{fig:loss-ablation} isolates the role of each loss on \texttt{Lift}. Removing embedding reconstruction causes the largest collapse, reducing both planning success and SSIM, which confirms that $\mathcal{L}_{\text{rec}}$ anchors the compact latent to the foundation embedding space. Removing proprioceptive supervision mostly preserves SSIM but lowers SR, showing that alignment shapes the latent for control rather than visual fidelity. The $\zz^s$/$\zz^c$ split dimension has only a modest effect, suggesting that performance is not sensitive to this hyperparameter.

\section{Conclusion}
In this paper, we propose {\method}, a task-sufficient world modeling framework that abstracts visual foundation embeddings into compact, task-centric latent spaces for accurate world modeling, planning, and control. By learning dynamics in this latent space rather than directly in high-dimensional embeddings, {\method} improves predictive accuracy and planning performance in reward-free offline settings. Our results highlight the importance of projecting visual foundation embeddings into a task-centric subspace for world modeling and effective control, rather than using them verbatim.

\paragraph{Limitations and Future Work.}
{\method} requires a training-time signal that exposes physical information (here proprioception $\ss^p_t$); when unavailable, distilled physical proxies or multi-view supervision are natural substitutes. Two follow-up directions we find promising are using {\method} as a lightweight downstream-alignment module on top of a larger pretrained generative world model, and training a generative dynamics model directly on the foundation embedding space to combine pixel-free supervision with control-centric structure. Details are in \Cref{app:future-work}.

\bibliographystyle{plainnat}
\bibliography{neurips2026}

\newpage
\appendix
\onecolumn

\begin{center}
\hrule height .5pt
\vspace{4mm}
{\Large\textbf{\textit{Appendix} of ``\mytitle''}}
\vspace{2mm}
\hrule height .5pt
\end{center}

\startcontents[sections]
{%
\hypersetup{linkcolor=black}%
\printcontents[sections]{l}{1}{\setcounter{tocdepth}{3}}%
}

\vspace{4mm}
\hrule height .5pt

\clearpage
\startcontents
\setcounter{figure}{0}
\renewcommand{\thefigure}{A\arabic{figure}}
\setcounter{table}{0}
\renewcommand{\thetable}{A\arabic{table}}
\renewcommand{\thesection}{\Alph{section}}
\setcounter{equation}{0}
\renewcommand{\theequation}{A\arabic{equation}}
\setcounter{theorem}{0}
\renewcommand{\thetheorem}{A\arabic{theorem}}
\setcounter{corollary}{0}
\renewcommand{\thecorollary}{A\arabic{corollary}}

\section{Experiment Details}

\subsection{Environments and Dataset Collection}
\label{sec:data-details}
We evaluate across nine environments. Below we describe the data source for each group, organized by collection method.

\paragraph{DINO-WM released data (\texttt{Maze}, \texttt{Wall}, \texttt{Push-T}).}
We directly adopt the released datasets from DINO-WM~\citep{zhou2025dinowm}.
\texttt{Maze} follows the PointMaze environment of D4RL~\citep{fu2020d4rl}, where a force-actuated 2-DoF ball navigates to a goal; the dataset contains 2000 random trajectories.
\texttt{Wall} is a 2D room-to-room navigation task with 1920 fixed-wall and 10240 randomized-wall trajectories.
\texttt{Push-T}, adapted from \citet{chi2025diffusion}, uses 18.5k expert demonstrations and 20k randomized trajectories for a block-pushing task.

\paragraph{RoboMimic official demonstrations (\texttt{Lift}, \texttt{Can}, \texttt{Square}).}
We use the official demonstration datasets released by the RoboMimic benchmark~\citep{mandlekar2021matters}.
All three tasks are simulated in \texttt{robosuite}~\citep{zhu2020robosuite} with MuJoCo, using a 7-DoF Franka Panda robot.
Demonstrations are collected via the RoboTurk teleoperation platform~\citep{mandlekar2018roboturk}, where human operators control the robot arm through a smartphone interface.
We use the Proficient-Human (PH) demonstrations, which are collected by a single skilled operator: 200 trajectories each for \texttt{Lift} (\texttt{robosuite Lift}), \texttt{Can} (\texttt{PickPlaceCan}), and \texttt{Square} (\texttt{NutAssemblySquare}).
Image observations are rendered from two simulated cameras (a front-facing \texttt{agentview} and a wrist-mounted \texttt{robot0\_eye\_in\_hand}).

\paragraph{TD-MPC2 rollouts (\texttt{Reacher}, \texttt{Cheetah}, \texttt{Hopper}).}
We generate offline trajectories by rolling out pretrained TD-MPC2 checkpoints from \citet{hansen2025learning} in the DeepMind Control Suite.
For each task we collect trajectories over multiple rounds, with actions produced by the TD-MPC2 policy conditioned on the current observation.
We record rendered frames and low-dimensional state observations at each step.
To ensure demonstration quality we apply a return-based filter: episodes are accepted only if their total return exceeds $0.75\times\text{median}$ (or $1.25\times\text{median}$ when negative).

\begin{table}[h]
\centering
\small
\caption{Action and proprioception dimensionalities of all nine evaluation environments. The \texttt{Robomimic} manipulation tasks (\texttt{Lift}, \texttt{Can}, \texttt{Square}) have markedly higher-dimensional action and proprioceptive spaces than the navigation and DMC continuous-control benchmarks, justifying our use of LDP and our claim that they form the most challenging planning regime in this study.}
\label{tab:dataset-dims}
\begin{tabular}{llccll}
\toprule
\textbf{Category} & \textbf{Dataset} & \textbf{Action dim} & \textbf{Proprio dim} & \textbf{Action type} & \textbf{Planner} \\
\midrule
\multirow{2}{*}{Navigation}
  & \texttt{Maze}    & 2 & 4  & continuous force $(F_x, F_y)$ & CEM \\
  & \texttt{Wall}    & 2 & 4  & continuous velocity $(v_x, v_y)$ & CEM \\
\midrule
Pushing
  & \texttt{Push-T}  & 2 & 5  & continuous 2D target $(t_x, t_y)$ & CEM \\
\midrule
\multirow{3}{*}{\textbf{Manipulation}}
  & \texttt{Lift}    & \textbf{7} & \textbf{43} & 6\,DoF joint velocity + gripper & LDP \\
  & \texttt{Can}     & \textbf{7} & \textbf{43} & 6\,DoF joint velocity + gripper & LDP \\
  & \texttt{Square}  & \textbf{7} & \textbf{43} & 6\,DoF joint velocity + gripper & LDP \\
\midrule
\multirow{3}{*}{DMC control}
  & \texttt{Reacher} & 2 & 6  & continuous joint torque & CEM \\
  & \texttt{Cheetah} & 6 & 17 & continuous joint torque & CEM \\
  & \texttt{Hopper}  & 4 & 15 & continuous joint torque & CEM \\
\bottomrule
\end{tabular}
\end{table}

\subsection{Details of Selected Visual Foundation Models}
\label{app:vis_fm}
\paragraph{DINOv2.}
We use the TorchHub implementation from \texttt{facebookresearch/dinov2} with backbone \texttt{dinov2\_vits14}.
Given an input image tensor $x \in \mathbb{R}^{B\times 3\times H\times W}$, we extract patch-level features as the representation.
This yields a sequence of normalized patch tokens of shape $(B, N, D)$, where $D=\texttt{num\_features}$ (384 for ViT-S/14)
and $N=(H/\texttt{patch\_size})\cdot(W/\texttt{patch\_size})$ with $\texttt{patch\_size}=14$, matching the configuration used by DINO-WM~\citep{zhou2025dinowm}.

\paragraph{DINOv3.}
We use the version \texttt{facebook/dinov3-vitb16-pretrain-lvd1689m}. The encoder takes RGB images
$x \in \mathbb{R}^{B \times 3 \times H \times W}$ as input, which are normalized using the model-provided
mean and standard deviation when available (otherwise ImageNet statistics).
DINOv3 employs a ViT-B/16 backbone with patch size $16\times16$ and hidden dimension $D=768$.
The raw output has shape
$(B, 1 + R + N, D)$, where $N=(H/16)\cdot(W/16)$ denotes the number of patch tokens and $R$ the number of register tokens.
Following standard practice, we discard the CLS token and any register tokens, retaining only the patch tokens of the shape $\mathbb{R}^{B \times N \times 768}$ as the final visual representation.

\paragraph{Cosmos CI Tokenizer.}
For Cosmos, we use the \texttt{NVIDIA/Cosmos-0.1-Tokenizer-CI8x8} tokenizer encoder loaded from a TorchScript checkpoint.
Images are resized to $112\times112$ and encoded in FP16. The encoder produces latents of shape $(B, C, H', W')$
(with $C=16$ in our setting), which we reshape into tokens $(B, H'W', C)$.
With patch size $8$ and image size $112$, this yields a $14\times14$ token grid (\ie, $196$ tokens), aligning with the token-count budget used by the rest of the architecture.

\subsection{Pipeline of Planning on Manipulation Tasks} \label{sec:plan_alg}
In this subsection, we describe the latent-space planning and control procedure. Depending on the task setup and data modality, we employ the Cross-Entropy Method (CEM) for navigation tasks and Latent Diffusion Planning (LDP) framework~\citep{xie2025latent} for manipulation tasks, based on the complexity of action space.
\paragraph{On Navigation Tasks.} For low-dimensional action spaces we plan directly with CEM, treating the encoded goal latent as the target state. 
\begin{enumerate}[leftmargin=*, label=(\arabic*)]

\item \textbf{Latent Encoding.}  
Given the current observation $\oo_0$ and the goal observation $\oo_g$, both represented as RGB images, we first extract visual embeddings $\xx_0=f_{\text{emb}}(\oo_0)$ and $\xx_g=f_{\text{emb}}(\oo_g)$, then encode them into latent states:
\[
\zz_0 = q(\zz_0 \mid \xx_0), \quad \zz_g = q(\zz_g \mid \xx_g).
\]

\item \textbf{Planning Objective.}  
The objective is to minimize the distance between the predicted final latent $\hat{\zz}_{H_p}$ and the goal latent $\zz_g$:
\[
\mathcal{C} = \|\hat{\zz}_{H_p} - \zz_g\|^2,
\quad \text{where} \quad
\hat{\zz}_{H_p} = f_\theta^{H_p}(\zz_0,\aa_{0:H_p-1}).
\]

\item \textbf{Action Sampling.}  
At each planning iteration, CEM samples a population of $N$ action sequences 
$\{\aa_{0:H_p-1}^i\}_{i=1}^N$ from a Gaussian distribution.  
The initial mean and covariance are initialized as $\boldsymbol{\mu}_0$ and $\boldsymbol{\Sigma}_0$.

\item \textbf{Trajectory Rollout.}  
For each sampled sequence, the learned world model predicts the latent rollout:
\[
\hat{\zz}_t = f_\theta^{t}(\zz_0,\aa_{0:t-1}), \quad t = 1, \dots, H_p,
\]
and the trajectory cost $\mathcal{C}^i$ is computed according to the objective above.

\item \textbf{Elite Selection and Update.}  
The top $K$ trajectories with the lowest costs are selected as elites, and the sampling distribution is updated by:
\[
\boldsymbol{\mu} \leftarrow \text{mean}(\aa^{(1:K)}), \quad 
\boldsymbol{\Sigma} \leftarrow \text{cov}(\aa^{(1:K)}).
\]

\item \textbf{Iteration.}  
A new population of $N$ trajectories is resampled from the updated Gaussian, and the process is repeated for a fixed number of iterations $T_{\text{cem}}$.

\item \textbf{Execution.}  
After optimization, the first $H_a$ actions $\aa_{0:H_a-1}$ are executed in the environment.  
At the next timestep, a new observation $\xx_{H_a}$ is encoded and the planning process repeats, forming a closed-loop receding-horizon controller.

\end{enumerate}

\paragraph{On Manipulation Tasks.} For high-dimensional, contact-rich actions we couple {\method} with the Latent Diffusion Planning (LDP) framework~\citep{xie2025latent}.
\begin{enumerate}[leftmargin=*, label=(\arabic*)]
    \item \textbf{Input Encoding.}  
    Given the current observation $\oo_0$, goal observation $\oo_g$, and proprioceptive state $\ss^p_0$, we use {\method} to encode observations into latent states:
    \[
    \zz_0 = q(\zz_0 \mid f_{\text{emb}}(\oo_0)), \quad \zz_g = q(\zz_g \mid f_{\text{emb}}(\oo_g)).
    \]

    \item \textbf{Planning Objective.}  
    The planner forecasts a sequence of future latent states $[\hat{\zz}_{k+1}, \dots, \hat{\zz}_{k+H_p}]$ conditioned on $\zz_k$, optimized by the diffusion loss:
    \[
    \mathcal{L}_{\text{planner}}(\psi) = \mathbb{E}_{t,\epsilon}\big[\|\epsilon_{\psi}(\hat{\zz}_{k+1}, \dots, \hat{\zz}_{k+H_p}; \zz_k, t) - \epsilon\|^2\big].
    \]

    \item \textbf{Inverse Dynamics Model.}  
    The IDM reconstructs actions between latent states via:
    \[
    \mathcal{L}_{\text{IDM}}(\xi) = \mathbb{E}_{t,\epsilon}\big[\|\epsilon_{\xi}(\hat{\aa}_k; \zz_k, \zz_{k+1}, t) - \epsilon\|^2\big].
    \]

    \item \textbf{Latent Trajectory Sampling.}  
    At inference, the planner samples $\hat{\zz}_{k+1}, \dots, \hat{\zz}_{k+H_p} \sim \mathcal{N}(0, I)$ and iteratively denoises via DDPM updates using $\epsilon_{\psi}$.

    \item \textbf{Action Reconstruction.}  
    For each pair $(\hat{\zz}_{k+i}, \hat{\zz}_{k+i+1})$, actions $\hat{\aa}_{k+i}$ are generated through diffusion updates with $\epsilon_{\xi}$.

    \item \textbf{Execution and Replanning.}  
    The first $H_a$ actions $\hat{\aa}_{k}, \dots, \hat{\aa}_{k+H_a-1}$ are executed, and the process repeats with new latent observation $\zz_{k+H_a}$, forming a closed-loop latent MPC scheme.
\end{enumerate}

\subsection{Implementation Details of Baselines}
\label{sec:imp_details}
\paragraph{Offline Version of DreamerV3}
We implement an offline variant of DreamerV3~\citep{hafner2023mastering} that trains exclusively from a fixed replay buffer without any environment interaction. The offline pipeline reads precomputed trajectories (images $\oo_t$, actions $\aa_t$, and optional proprio/state and rewards), builds the DreamerV3 replay, and runs the standard DreamerV3 world‑model training loop on randomly sampled batches. Evaluation is performed on held‑out trajectories by reconstructing and open‑loop predicting image sequences: the model encodes a context window of observed frames (posterior) and imagines the future (prior), after which the decoder produces reconstructions/predictions that are compared to ground truth. We keep the original DreamerV3 architecture and losses, and only adapt the data loading, replay ingestion, and offline evaluation logic to accommodate reward‑free datasets and dataset‑specific observation formats.

\paragraph{Offline Version of TD-MPC2} We implement TD-MPC2‑Offline by adding a replay‑only training pipeline to the official TD‑MPC2 codebase~\citep{hansen2024tdmpc}. The offline trainer removes all environment interaction and learns solely from a fixed dataset. \texttt{Robomimic} demonstrations are loaded from preconverted episode files (RGB video, actions, rewards, terminals), and a fixed replay buffer is built once at startup. Training samples random sequences from this buffer, matching the original TD‑MPC2 update equations; no online data is added. For evaluation, we introduce an offline validation split and compute losses on held‑out sequences without parameter updates. To align with visual world‑model diagnostics, we add a reconstruction decoder and rollout visualization: the decoder is trained with reconstruction loss (with backpropagating into the latent), and rollouts are generated by iterating the latent predictor and decoding predicted latents. Rollout visualization uses the same format as our method. We also extend rollout logic to respect frame skipping by applying the full action sequence within each skip window (rather than subsampling actions), ensuring motion fidelity in long‑horizon rollouts.

\paragraph{Offline Version of MuZero}\label{app:muzero-offline}
We adapt MuZero~\citep{schrittwieser2020mastering} to the offline setting by keeping its three-network decomposition---representation $h$, dynamics $g$, and prediction $f$---and supervising the joint model on fixed trajectories instead of MCTS-driven self-play. In our implementation, $h$ is the frozen DINOv2 encoder followed by the same compact linear projector as {\method}, $g$ is a Transformer next-latent predictor that consumes $(\zz_t,\aa_t)$, and $f$ collects the per-step heads over the latent. As our datasets carry no dense reward, $f$ reduces to a policy head $f^{\pi}_{\xi}$ trained by behaviour cloning on the recorded action; the reward and value heads remain available in the codebase and are deactivated only because the supervision is missing. The training loss is $\mathcal{L}_{\text{cons}} + 0.25\,\|f^{\pi}_{\xi}(\zz_t)-\aa_t\|_2^2$, where $\mathcal{L}_{\text{cons}}$ matches the unrolled latent to the encoded next-step latent with a stop-gradient on the target, as in the original MuZero. With reward labels the standard $\mathcal{L}_{\text{cons}} + \mathcal{L}_{r} + 0.25\,\mathcal{L}_{v} + 0.25\,\mathcal{L}_{\pi}$ objective is recovered by simply enabling the reward and Monte-Carlo value heads.

\subsection{Compute Resources}\label{app:compute}
All training and evaluation runs use a single NVIDIA H100 80\,GB GPU per experiment.
A typical {\method} run on \texttt{Lift} / \texttt{Can} / \texttt{Square} converges in roughly $8$ hours over $100$ epochs at peak memory $\approx 35$\,GB; \texttt{Push-T} / \texttt{Wall} / \texttt{Maze} converge in $1$--$2$ hours; the DMC tasks (\texttt{Reacher}, \texttt{Cheetah}, \texttt{Hopper}) take $4$--$6$ hours each. Each baseline (TD-MPC2, DreamerV3, MuZero, DINO-WM) uses the same single-H100 budget per environment, with comparable wall-clock times.
Aggregating training, evaluation, and the linear-probe / unseen-perturbation studies, the experiments reported in the paper consume roughly $1{,}200$ H100 GPU-hours.

\begin{table}[t]

\centering

\caption{Physical interpretation of state observations and their partitioning into three task-centric variable groups for the \texttt{Robomimic} datasets, based on their roles in the robot trajectory. \textcolor{gray}{Gray} entries are \emph{not} used during training; they serve only as ground-truth references for planning.}

\label{tab:robomimic-states}

\footnotesize

\setlength{\tabcolsep}{3pt}

\renewcommand{\arraystretch}{1.15}

\resizebox{\textwidth}{!}{%

\begin{tabular}{@{}l p{6.5cm} p{3.5cm} p{3.7cm}@{}}

\toprule

\textbf{Dataset} & \textbf{Proprio} & \textbf{Object} & \textbf{Goal} \\

\midrule
\texttt{Lift} &
\texttt{robot0\_joint\_pos}, \texttt{robot0\_joint\_pos\_sin}, \texttt{robot0\_joint\_pos\_cos}, \texttt{robot0\_joint\_vel}, \texttt{robot0\_eef\_pos}, \texttt{robot0\_eef\_quat}, \texttt{robot0\_eef\_quat\_site}, \texttt{robot0\_gripper\_qpos}, \texttt{robot0\_gripper\_qvel}
&
\textcolor{gray}{\texttt{object\_pos}, \texttt{object\_quat}}
&
\textcolor{gray}{\texttt{grip\_obj\_rel\_vec}} \\
\addlinespace[2pt]
\texttt{Can} &
\texttt{robot0\_joint\_pos}, \texttt{robot0\_joint\_pos\_sin}, \texttt{robot0\_joint\_pos\_cos}, \texttt{robot0\_joint\_vel}, \texttt{robot0\_eef\_pos}, \texttt{robot0\_eef\_quat}, \texttt{robot0\_eef\_quat\_site}, \texttt{robot0\_gripper\_qpos}, \texttt{robot0\_gripper\_qvel}
&
\textcolor{gray}{\texttt{object\_pos}, \texttt{object\_quat}}
&
\textcolor{gray}{\texttt{target\_pos}, \texttt{target\_quat}} \\
\addlinespace[2pt]
\texttt{Square} &
\texttt{robot0\_joint\_pos}, \texttt{robot0\_joint\_pos\_sin}, \texttt{robot0\_joint\_pos\_cos}, \texttt{robot0\_joint\_vel}, \texttt{robot0\_eef\_pos}, \texttt{robot0\_eef\_quat}, \texttt{robot0\_eef\_quat\_site}, \texttt{robot0\_gripper\_qpos}, \texttt{robot0\_gripper\_qvel}
&
\textcolor{gray}{\texttt{object\_pos}, \texttt{object\_quat}}
&
\textcolor{gray}{\texttt{target\_pos}, \texttt{target\_quat}} \\
\bottomrule
\end{tabular}
}
\end{table}


\section{Theoretical Rationale: When and How can We Learn Task-centric Representation from Visual Foundations?}
\label{sec:thm}

This appendix expands the theoretical statement of \Cref{thm:tcwm-main}. \Cref{app:thm-illustration} discusses the motivation, the underlying hierarchical generative model, the assumptions, and an empirical sanity check on \texttt{Lift}. \Cref{app:thm-proof} contains the full proof.

\subsection{Illustration and Analysis}\label{app:thm-illustration}

Understanding when and how task-centric representation can be learned from visual foundation models is critical for modern world-model training. We study this problem through \textit{identifiability}, the asymptotic guarantee of \textit{recovering} a semantically meaningful latent space. The discussion is organized around three previously underexplored questions: \texttt{(1)} how imposing a latent Markov structure on top of visual foundations helps guide world representations; \texttt{(2)} why a simple linear projection is sufficient to transform rich visual features into task-relevant world representations; and \texttt{(3)} empirically validating these claims through ablation studies and linear probing experiments.

We first establish identifiability of task-centric representation in a general framework.

\begin{definition}[Recoverability of Task-Centric Representation]
The estimated $\hzz^s_t$ is a nontrivial transformation of the true $\zz^s_t$ in the defined process of Sec.~\ref{sec:overall-model} for each frame.
\end{definition}
We next show that, under mild regularity assumptions, our formulation described in Sec.~\ref{sec:overall-model} guarantees such recoverability.

\begin{figure}[H]
\centering
\begin{tikzpicture}[
    >=Stealth, font=\small,
    latent/.style={circle, draw=black!70, thick, minimum size=0.7cm, inner sep=0pt},
    obs/.style={rectangle, draw=black!70, thick, minimum size=0.7cm, inner sep=2pt, fill=gray!12},
    arr/.style={draw=black!60, thick, -{Stealth[length=1.8mm]}},
    tarr/.style={draw=blue!50!cyan!70!black, thick, -{Stealth[length=1.8mm]}},
    harr/.style={draw=red!60!black, thick, -{Stealth[length=1.8mm]}},
    note/.style={font=\scriptsize, text=gray!50!black},
    zbox/.style={draw=orange!50, dashed, rounded corners, inner sep=4pt, fill=orange!6},
]

\node[note, font=\footnotesize] at (0,5.05) {$t{-}1$};
\node[note, font=\footnotesize] at (2.5,5.05) {$t$};
\node[note, font=\footnotesize] at (5.0,5.05) {$t{+}1$};

\node[note, font=\footnotesize\bfseries, text=orange!70!black] at (-2.0,3.7) {Layer 2};
\node[latent, fill=orange!15, draw=orange!60] (zs0) at (0,4.2) {$\zz^s$};
\node[latent, fill=gray!10] (zc0) at (0,3.2) {$\zz^c$};
\node[latent, fill=orange!15, draw=orange!60] (zs1) at (2.5,4.2) {$\zz^s$};
\node[latent, fill=gray!10] (zc1) at (2.5,3.2) {$\zz^c$};
\node[latent, fill=orange!15, draw=orange!60] (zs2) at (5.0,4.2) {$\zz^s$};
\node[latent, fill=gray!10] (zc2) at (5.0,3.2) {$\zz^c$};

\begin{scope}[on background layer]
\node[zbox, fit=(zs0)(zc0)] (zbox0) {};
\node[zbox, fit=(zs1)(zc1)] (zbox1) {};
\node[zbox, fit=(zs2)(zc2)] (zbox2) {};
\end{scope}

\draw[tarr] (zbox0.east) -- (zbox1.west);
\draw[tarr] (zbox1.east) -- (zbox2.west);

\node[note, font=\footnotesize\bfseries, text=blue!50] at (-2.0,2.0) {Layer 1};
\node[latent, fill=blue!10, draw=blue!50] (x0) at (0,2.0) {$\xx$};
\node[latent, fill=blue!10, draw=blue!50] (x1) at (2.5,2.0) {$\xx$};
\node[latent, fill=blue!10, draw=blue!50] (x2) at (5.0,2.0) {$\xx$};

\draw[tarr, blue!30, dashed] (x0) -- (x1);
\draw[tarr, blue!30, dashed] (x1) -- (x2);

\node[note, font=\footnotesize\bfseries, text=gray!50] at (-2.0,0.5) {Layer 0};
\node[obs] (o0) at (0,0.5) {$\oo$};
\node[obs] (o1) at (2.5,0.5) {$\oo$};
\node[obs] (o2) at (5.0,0.5) {$\oo$};

\draw[arr] (o0) -- (x0);
\draw[arr] (o1) -- node[left, note] {$f_{\text{enc}}$} (x1);
\draw[arr] (o2) -- (x2);

\draw[arr] (x0) -- (zbox0.south);
\draw[arr] (x1) -- (zbox1.south);
\draw[arr] (x2) -- (zbox2.south);

\draw[arr, dashed, gray!55] (zbox1.south east) to[bend left=40] node[right, note, text=gray!55!black, pos=0.55] {$\mathcal{L}_{\text{rec}}$} (x1.north east);

\node[obs, fill=red!8, draw=red!50] (sp) at (2.5,5.9) {$\ss^p$};
\draw[harr, dashed] (sp) -- node[right, note] {$\mathcal{L}_{\text{align}}$} (zs1);

\draw[thick, gray!30, decorate, decoration={brace, amplitude=4pt}] (-1.2,1.6) -- (-1.2,3.6);
\node[note, text=blue!60!black, rotate=90, anchor=south, font=\tiny] at (-1.6,2.6) {A1, A3};

\draw[thick, gray!30, decorate, decoration={brace, amplitude=4pt}] (-1.2,0.1) -- (-1.2,1.5);
\node[note, text=blue!60!black, rotate=90, anchor=south, font=\tiny] at (-1.6,0.8) {A2};

\draw[thick, gray!30, decorate, decoration={brace, amplitude=4pt, mirror}] (3.6,4.7) -- (3.6,6.1);
\node[note, text=red!60!black, font=\tiny] at (4.1,5.4) {A4};

\node[note, align=left] at (6.8,3.7) {\shortstack[l]{Task-centric\\latents}};
\node[note, align=left] at (6.8,2.0) {\shortstack[l]{FM embeddings\\\Snowflake\ frozen}};
\node[note, align=left] at (6.8,0.5) {Observations};

\end{tikzpicture}
\caption{Graphical model of TC-WM's hierarchical structure. Circles denote latent variables; squares denote observed variables. Observations $\oo_t$ are mapped to frozen FM embeddings $\xx_t$ (Layer~1), then projected to a compact latent $\zz_t = [\zz^s_t, \zz^c_t]$ (Layer~2, dashed orange box). The unified temporal arrows act on the joint latent $\zz_t$. $\mathcal{L}_{\text{align}}$ aligns $\zz^s_t$ with proprioception $\ss^p_t$; $\mathcal{L}_{\text{rec}}$ (curved) regularises by reconstructing the embedding from $\zz_t$. Assumptions A1--A4 of Theorem~\ref{thm:blk idn} are annotated.}
\label{fig:hierarchical}
\end{figure}
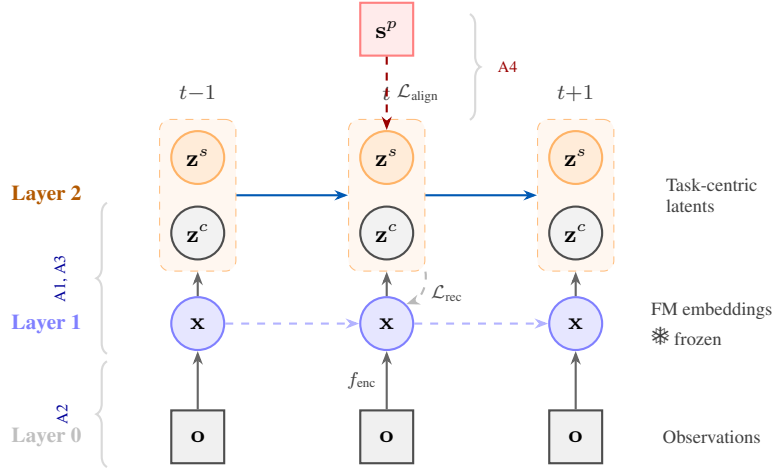

\paragraph{A View from Hierarchical Model.}
To justify that learning dynamics from frozen visual embeddings is theoretically sound, we show that the underlying task-centric world representation $\hzz_t^s$ faithfully captures the physical factors governing the environment. As illustrated in Figure~\ref{fig:hierarchical}, our architecture forms a three-level hierarchy: raw observations $\oo_t$ are mapped to foundation embeddings $\xx_t$ by a frozen encoder, which are then projected into a compact latent $\zz_t = [\zz^s_t, \zz^c_t]$. The theorem below guarantees that this hierarchical process preserves the task-relevant factors.

We first formalize this property through the concept of \emph{recoverability}, which ensures that the learned latents are meaningful transformations of the true task-relevant variables.

\clearpage
\refstepcounter{theorem}\label{thm:blk idn}
\begin{tcwmtheorembox}{Theorem~\thetheorem\ \textbf{\textit{(Task-centric identifiability from visual embeddings)}}}
\textit{Suppose images $\oo_t$, frozen embeddings $\xx_t$, and latent variables $\zz_t=[\zz^s_t,\zz^c_t]$ follow the process in~\Cref{sec:overall-model}, and the learned latent world model matches the true joint distribution of adjacent embeddings $\{\xx_{t-1},\xx_t,\xx_{t+1}\}$. Assume:}
\begin{description}[leftmargin=0pt, labelwidth=2.7em, labelsep=0.55em, style=sameline, itemsep=0.2em, topsep=0.2em]
    \item[\textbf{A1.}] \textbf{\textit{(\underline{Contextual dynamics}):}} \textit{$L_{\xx_{t+1:t+H}\mid \zz_t}$ and $L_{\xx_{t-H:t-1}\mid \xx_{t+1:t+H}}$ are injective and bounded.}
    \item[\textbf{A2.}] \textbf{\textit{(\underline{Latent--observed variability}):}} \textit{For any $\zz_t^{(1)}\neq \zz_t^{(2)}$, $p(\xx_t\mid \zz_t^{(1)})\neq p(\xx_t\mid \zz_t^{(2)})$.}
    \item[\textbf{A3.}] \textbf{\textit{(\underline{Differentiability}):}} \textit{There exists a differentiable $F$ such that $F[p_{\xx_t\mid\zz_t}(\cdot\mid\zz_t)]=h_z(\zz_t)$.}
    \item[\textbf{A4.}] \textbf{\textit{(\underline{Partial alignment}):}} \textit{The alignment objective in~\Cref{eq:infonce} is sufficiently minimized.}
\end{description}
\textit{Then the learned $\zz_t$ is block-wise identifiable, and the task-centric $\zz^s_t$ is affine-identifiable:}
\begin{equation*}
\resizebox{\linewidth}{!}{$
\underbrace{
\hzz_t = h_z(\zz_t),\quad h_z \text{ invertible and } C^1
}_{\substack{\text{\footnotesize \textbf{Latent Space Recovery:} the compact latent preserves}\\[-0.15ex]\text{\footnotesize the true world state up to reparameterization}}}
\qquad
\underbrace{
\zz^s_t = \mathbf{A}\hzz^s_t + \mathbf{b}
}_{\substack{\text{\footnotesize \textbf{Task-centric recovery:} the aligned block recovers}\\[-0.15ex]\text{\footnotesize physical factors up to an affine map}}}
$}
\end{equation*}
\end{tcwmtheorembox}
\begin{tcwmtheorembox}{Takeaway}
This theorem justifies our hierarchical paradigm: even though {\method} learns dynamics in the foundation-embedding space $\xx_t$ rather than in raw pixels $\oo_t$, the reconstruction and alignment objectives suffice to recover the physically meaningful latent factors that drive the environment. The result connects our approach to a broader family of ``dynamics on pretrained encoders,'' including recent generative and latent diffusion frameworks such as VideoLDM~\citep{blattmann2023align}, DiT~\citep{peebles2023scalable}, RAW~\citep{zheng2025diffusion}, and REPA~\citep{yu2025representation}.
\end{tcwmtheorembox}

\paragraph{Proof Sketch.}
Our argument unfolds in three steps.
\texttt{(1)} Under nonparametric assumptions, embeddings $\xx_t$ are invertible transformations of images $\oo_t$ (\eg, DINO or $\beta$-VAE), thus preserving the environment's essential factors.
\texttt{(2)} For the embedding-level latent $\zz_t$, we construct an integral operator over $2H{+}1$ adjacent observations and show, via Markov conditional independence and a spectral decomposition argument, that any model matching the true joint distribution of $\{\xx_{t-H:t+H}\}$ recovers $\zz_t$ up to an invertible reparametrisation $h_z$.
\texttt{(3)} For the task-centric subspace $\zz^s_t$, the InfoNCE alignment with the proprioception $\ss^p_t$ collapses $h_z$ on the $\zz^s$-block to a map that depends only on $\zz^s_t$, and an exponential-family/dot-product argument then forces it to be affine.
The full proof is given below.

\paragraph{Empirical Verification of Assumptions on \texttt{Lift}.}
We complement the theory with direct empirical checks of A1--A4 on a fully trained {\method} checkpoint (\texttt{Lift}, $d_z = 256$, $d_s = 64$). All probes operate on the patch-mean-pooled latent $\zz_t$ and the mean-pooled foundation embedding $\xx_t$; all randomness uses fixed seeds.

\begin{figure}[h]
    \centering
    \includegraphics[width=0.96\linewidth]{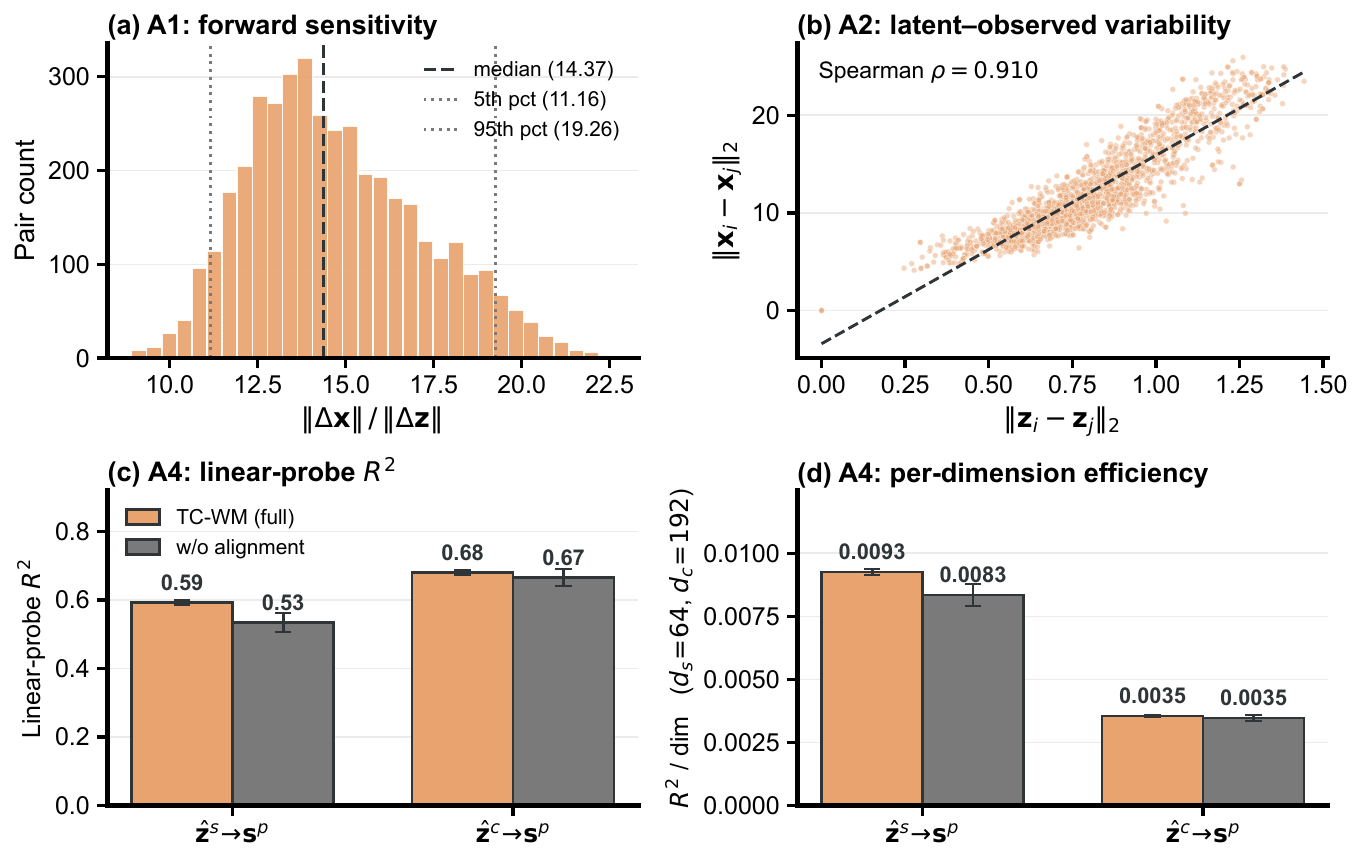}
    \caption{Empirical verification of assumptions A1, A2, and A4 on \texttt{Lift}.
    \textbf{(a)} A1 (Contextual Dynamics): distribution of the forward-sensitivity ratio $\|\Delta\xx\|/\|\Delta\zz\|$ over $4096$ validation pairs --- the ratio is bounded above and away from zero, supporting injectivity and boundedness of the context operator.
    \textbf{(b)} A2 (Latent--Observed Variability): scatter of $\|\zz_i-\zz_j\|$ against $\|\xx_i-\xx_j\|$ for $2037$ validation pairs (Spearman $\rho=0.910$) --- distinct latents map to distinguishable embedding distributions.
    \textbf{(c)} A4 (Partial Alignment): linear-probe $R^2$ from $\hzz^s_t$ and $\hzz^c_t$ to $\ss^p_t$ (5-fold CV, ridge $\alpha{=}1$); the InfoNCE alignment lifts $R^2(\hzz^s\!\to\!\ss^p)$ from $0.534$ to $0.592$ while leaving $\hzz^c$ essentially unchanged.
    \textbf{(d)} The same comparison on a per-dimension basis ($R^2$ divided by $d_s{=}64$ or $d_c{=}192$): the task-centric block is more than $2\times$ as efficient at encoding proprioception, confirming that the alignment concentrates proprioceptive information in $\zz^s$.}
    \label{fig:asm_verify}
\end{figure}

\textbf{A1 (Contextual Dynamics).}
A1 demands that $L_{\xx_{t+1:t+H}\mid \zz_t}$ be both bounded and injective. We probe the closely related forward sensitivity $\|\Delta\xx\|/\|\Delta\zz\|$ on $4096$ random pairs of (latent, embedding) co-drawn from validation trajectories. The ratio is tightly concentrated, with median $14.37$, $5$th percentile $11.16$, and $95$th percentile $19.26$ (\Cref{fig:asm_verify}); the lower tail is bounded away from zero, evidencing injectivity, and the upper tail is bounded, evidencing the operator-norm part of A1.

\textbf{A2 (Latent--Observed Variability).}
A2 demands that distinct latents yield distinct conditional embeddings, $\zz^{(1)}_t \neq \zz^{(2)}_t \Rightarrow p(\xx_t\mid\zz^{(1)}_t) \neq p(\xx_t\mid\zz^{(2)}_t)$. As a population-level surrogate we measure how strongly latent distance predicts embedding distance: \Cref{fig:asm_verify} plots $\|\zz_i - \zz_j\|$ against $\|\xx_i - \xx_j\|$ over $2037$ random validation pairs. The two distance series are nearly comonotone, with Spearman $\rho = 0.910$ and Pearson $\rho = 0.911$; in particular, no points cluster near the $\Delta\xx \approx 0$ axis at non-zero $\Delta\zz$, ruling out the failure mode that A2 forbids.

\textbf{A3 (Differentiability).}
A3 is satisfied by construction: the projector and embedding decoder are smooth $C^1$ neural networks, the conditional density $p(\xx_t\mid\zz_t)$ is the output of a differentiable Gaussian decoder, and the functional $F$ of A3 can be taken as a moment-matching label, which is differentiable in $\zz_t$ via the implicit function theorem. No further empirical check is needed.

\textbf{A4 (Partial Alignment).}
A4 demands that the InfoNCE alignment loss be sufficiently minimized. We linearly probe both subspaces against the proprioceptive state $\ss^p_t$ (\Cref{fig:asm_verify}). On the full {\method} checkpoint the task-centric block achieves $R^2(\hzz^s_t \to \ss^p_t) = 0.592 \pm 0.007$, a $+5.8$ absolute-point improvement over the alignment-free ablation ($0.534 \pm 0.027$). The complementary block $\hzz^c_t$ predicts $\ss^p_t$ at $R^2 = 0.680 \pm 0.008$, but its dimensionality is three times higher ($d_c = 192$ vs $d_s = 64$); on a per-dimension basis the task-centric block is more than twice as efficient at encoding proprioception ($0.0093$ vs $0.0035$ $R^2$ per dim), confirming that the InfoNCE objective concentrates the proprioceptive content in the designated $\zz^s$-block.

\subsection{Proof of Theorem~\ref{thm:blk idn}}\label{app:thm-proof}
\label{app:proof}

\paragraph{Notation and setup.}
We restate the data-generating process from \Cref{sec:overall-model} in a form convenient for the proof. Observations $\oo_t \in \mathcal{O}$ are mapped by a frozen visual foundation encoder $f_{\text{enc}}$ to embeddings $\xx_t = f_{\text{enc}}(\oo_t) \in \mathcal{X} \subseteq \mathbb{R}^{d_x}$, and the embedding is generated from a compact latent $\zz_t \in \mathcal{Z} \subseteq \mathbb{R}^{d_z}$ through a (possibly noisy) mixing
\begin{equation}
    \xx_t \,=\, g(\zz_t,\, \boldsymbol{\epsilon}^0_t),
    \qquad
    \boldsymbol{\epsilon}^0_t \,\perp\, \boldsymbol{\epsilon}^0_{t'}\ \text{for}\ t \neq t',
    \qquad
    \boldsymbol{\epsilon}^0_t \,\perp\, \zz_{1:T},
    \label{eq:gen-x}
\end{equation}
with the noise sequence $\{\boldsymbol{\epsilon}^0_t\}$ exogenous and i.i.d. The latent splits as $\zz_t = [\zz^s_t, \zz^c_t]$ with $\zz^s_t \in \mathcal{Z}^s \subseteq \mathbb{R}^{d_s}$ and $\zz^c_t \in \mathcal{Z}^c \subseteq \mathbb{R}^{d_z - d_s}$, and the latent process $\{\zz_t\}$ is first-order Markov,
\begin{equation}
    p(\zz_{t+1} \mid \zz_{1:t}) \,=\, p(\zz_{t+1} \mid \zz_t).
    \label{eq:gen-z}
\end{equation}
The proprioception used by the InfoNCE alignment of \Cref{eq:infonce} is a deterministic function of the task-centric block,
\begin{equation}
    \ss^p_t \,=\, m(\zz^s_t),
    \qquad
    m: \mathcal{Z}^s \to \mathcal{S}^p\ \text{is a $C^1$-diffeomorphism onto its image.}
    \label{eq:gen-sp}
\end{equation}
For any two random variables $\boldsymbol{u} \in \mathcal{U}$ and $\boldsymbol{v} \in \mathcal{V}$ with conditional density $p_{\boldsymbol{v}\mid\boldsymbol{u}}$, the induced \emph{linear (integral) operator} is
\begin{equation*}
    L_{\boldsymbol{v}\mid\boldsymbol{u}}\,:\, \mathcal{F}(\mathcal{U}) \;\longrightarrow\; \mathcal{F}(\mathcal{V}),
    \qquad
    [L_{\boldsymbol{v}\mid\boldsymbol{u}}\phi](\boldsymbol{v})
    \,:=\, \int_{\mathcal{U}} p(\boldsymbol{v} \mid \boldsymbol{u})\,\phi(\boldsymbol{u})\,\mathrm{d}\boldsymbol{u},
\end{equation*}
where $\mathcal{F}(\cdot)$ denotes a suitable space of $L^1$-densities, and is bounded whenever the conditional density is essentially bounded. We additionally use the multiplication operator $\Delta_{\xx_t}: \mathcal{F}(\mathcal{Z}) \to \mathcal{F}(\mathcal{Z})$ defined by $[\Delta_{\xx_t}\phi](\zz_t) := p(\xx_t \mid \zz_t)\,\phi(\zz_t)$, which acts as a (functional-valued) diagonal scaling indexed by $\zz_t$ and parametrised by $\xx_t$.

We use $\hat{\,\cdot\,}$ to denote estimated quantities: $\hzz_t \in \hat{\mathcal{Z}}$, $\hat g$, $\hat p$, etc. The hypothesis of Theorem~\ref{thm:blk idn} asserts that the estimated and true joint distributions of $\{\xx_{t-1}, \xx_t, \xx_{t+1}\}$ coincide.

\medskip\noindent
\textbf{Conditional independence structure.} The combination of \eqref{eq:gen-x}, the i.i.d.\ noise assumption, and the Markov property \eqref{eq:gen-z} implies the two relations
\begin{equation}
    \xx_{t-H:t-1}\;\perp\;\xx_{t+1:t+H}\,\bigm|\,\zz_t,
    \qquad
    \xx_t\;\perp\;\xx_{t-H:t-1},\,\xx_{t+1:t+H}\,\bigm|\,\zz_t,
    \tag{CI}\label{eq:ci}
\end{equation}
which we use throughout. The first relation says that, conditioned on the present latent $\zz_t$, past and future embeddings are independent; this is the Markov screening property combined with the fact that each $\xx_\tau$ is a noisy function of $\zz_\tau$ alone. The second says that the present embedding $\xx_t$ depends on the past and future only through $\zz_t$, which follows directly from~\eqref{eq:gen-x}: $\xx_t$ is a function of $(\zz_t, \boldsymbol{\epsilon}^0_t)$, and $\boldsymbol{\epsilon}^0_t$ is independent of every other quantity.

We first establish block-wise identifiability of $\zz_t$ from the matched joint distribution of $2H{+}1$ adjacent embeddings, and then refine this to the affine recovery on the task-centric subspace using the InfoNCE alignment objective.

Fix a time index $t$. The goal is to show that any estimated model whose joint distribution over $\{\xx_{t-H},\ldots,\xx_{t+H}\}$ matches the true one must agree with the true generative model up to an invertible reparametrisation of the latent at $t$. By the first relation in~\eqref{eq:ci}, the joint density of past and future windows admits the latent-mediated decomposition
\begin{equation}
    p(\xx_{t-H:t-1},\,\xx_{t+1:t+H})
    \,=\, \int_{\mathcal{Z}} p(\xx_{t-H:t-1} \mid \zz_t)\, p(\xx_{t+1:t+H} \mid \zz_t)\, p(\zz_t)\,\mathrm{d}\zz_t.
    \label{eq:joint-past-future}
\end{equation}
Dividing by the marginal $p(\xx_{t+1:t+H})$ and recognising $p(\zz_t \mid \xx_{t+1:t+H}) = p(\xx_{t+1:t+H} \mid \zz_t)\,p(\zz_t)/p(\xx_{t+1:t+H})$ on the right, this rewrites in operator form as
\begin{equation}
    L_{\xx_{t-H:t-1} \,\mid\, \xx_{t+1:t+H}}
    \,=\, L_{\xx_{t-H:t-1} \,\mid\, \zz_t}\,\circ\,L_{\zz_t \,\mid\, \xx_{t+1:t+H}}.
    \label{eq:op-factor}
\end{equation}
Each composing operator is bounded by A1. To see that $L_{\zz_t \mid \xx_{t+1:t+H}}$ is well defined and injective, observe that Bayes' rule yields
$L_{\zz_t \mid \xx_{t+1:t+H}} = D_{p(\zz_t)} \circ L_{\xx_{t+1:t+H} \mid \zz_t}^{*} \circ D_{p(\xx_{t+1:t+H})}^{-1}$,
where $D_q$ denotes the multiplication operator by density $q$ and $L^*$ is the formal adjoint; injectivity of $L_{\xx_{t+1:t+H}\mid\zz_t}$ from A1 transfers, via the adjoint and the strict positivity of the marginals on the relevant supports, to injectivity of the posterior operator on the range that arises in~\eqref{eq:op-factor}.\footnote{The strict positivity of the marginals on supports is implicit in the assumption that the data process is non-degenerate, as is standard in identifiability arguments for nonlinear ICA. The integral operators are taken on densities supported within $\mathcal{X}$ and $\mathcal{Z}$.} Combined with the injectivity of the cross-operator $L_{\xx_{t-H:t-1}\mid\xx_{t+1:t+H}}$ from A1, equation~\eqref{eq:op-factor} admits a unique left-inverse on the relevant function space; we will use this to invert the composition in Step~3.

We now bring the present embedding $\xx_t$ into the picture. By the second relation in~\eqref{eq:ci}, the four-way joint factorises with $\xx_t$ entering only through $p(\xx_t \mid \zz_t)$:
\begin{equation*}
    p(\xx_{t-H:t-1},\,\xx_t,\,\xx_{t+1:t+H})
    \,=\, \int_{\mathcal{Z}} p(\xx_{t-H:t-1} \mid \zz_t)\, p(\xx_t \mid \zz_t)\, p(\xx_{t+1:t+H} \mid \zz_t)\, p(\zz_t)\,\mathrm{d}\zz_t.
\end{equation*}
Dividing by the marginal $p(\xx_{t+1:t+H})$ and treating $\xx_t$ as a parameter (so that $p(\xx_t \mid \zz_t)$ becomes a $\zz_t$-indexed scalar), this rewrites as the operator identity
\begin{equation}
    L_{\xx_{t-H:t-1},\,\xx_t \,\mid\, \xx_{t+1:t+H}}
    \,=\, L_{\xx_{t-H:t-1} \mid \zz_t}\,\circ\,\Delta_{\xx_t}\,\circ\,L_{\zz_t \,\mid\, \xx_{t+1:t+H}}.
    \label{eq:op-with-xt}
\end{equation}
Equations~\eqref{eq:op-factor} and~\eqref{eq:op-with-xt} together yield
\begin{equation}
    \underbrace{L_{\xx_{t-H:t-1},\,\xx_t \,\mid\, \xx_{t+1:t+H}}\;\circ\;L^{-1}_{\xx_{t-H:t-1} \,\mid\, \xx_{t+1:t+H}}}_{=:\,T_{\xx_t}\,\text{(observable)}}
    \,=\, L_{\xx_{t-H:t-1} \mid \zz_t}\,\circ\,\Delta_{\xx_t}\,\circ\,L^{-1}_{\xx_{t-H:t-1} \mid \zz_t}.
    \label{eq:Tx-eigendecomp}
\end{equation}
Two facts about~\eqref{eq:Tx-eigendecomp} are central. First, the right-hand side is a similarity transform: $T_{\xx_t}$ is $\Delta_{\xx_t}$ written in the basis given by $L_{\xx_{t-H:t-1}\mid\zz_t}$. Since $\Delta_{\xx_t}$ is the diagonal multiplication operator with eigenvalue $p(\xx_t \mid \zz_t)$ at index $\zz_t$, the family $\{T_{\xx_t}\}_{\xx_t \in \mathcal{X}}$ is simultaneously diagonalised by the same eigenfunction system, namely the columns of $L_{\xx_{t-H:t-1} \mid \zz_t}$, with eigenvalues $\{p(\xx_t \mid \zz_t)\}_{\zz_t \in \mathcal{Z}}$. Second, $T_{\xx_t}$ is fully determined by the joint distribution of $\{\xx_{t-H:t-1}, \xx_t, \xx_{t+1:t+H}\}$, a quantity matched by the estimated model under the hypothesis of the theorem. Any estimated decomposition
\begin{equation*}
    T_{\xx_t}
    \,=\, \hat L_{\xx_{t-H:t-1} \mid \hzz_t}\,\circ\,\hat\Delta_{\xx_t}\,\circ\,\hat L^{-1}_{\xx_{t-H:t-1} \mid \hzz_t}
\end{equation*}
must therefore agree with the true one as an operator equality.

It remains to show that the simultaneous diagonalisation of the family $\{T_{\xx_t}\}_{\xx_t \in \mathcal{X}}$ uniquely determines the eigenvalue assignment $\zz_t \mapsto p(\xx_t \mid \zz_t)$ and the eigenfunction system $\zz_t \mapsto p(\xx_{t-H:t-1} \mid \zz_t)$, up to a single relabelling of the index set $\mathcal{Z}$.

We first use A2 to argue distinct eigenvalues. Suppose, towards a contradiction, that for some pair $\zz^{(1)}_t \neq \zz^{(2)}_t$ in $\mathcal{Z}$ the maps $\xx_t \mapsto p(\xx_t \mid \zz^{(1)}_t)$ and $\xx_t \mapsto p(\xx_t \mid \zz^{(2)}_t)$ coincide as functions on $\mathcal{X}$. This is exactly the negation of A2, and is excluded; consequently for almost every $\xx_t \in \mathcal{X}$ the eigenvalue assignment $\zz_t \mapsto p(\xx_t \mid \zz_t)$ is injective on $\mathcal{Z}$.

Next we leverage the simultaneous diagonalisation. Fix $\xx_t \in \mathcal{X}$ and let $\lambda(\zz_t) := p(\xx_t \mid \zz_t)$. The eigenvalues of $T_{\xx_t}$ are $\{\lambda(\zz_t)\}_{\zz_t}$ with associated eigenfunctions $\phi_{\zz_t}(\cdot) := p(\xx_{t-H:t-1} \mid \zz_t)(\cdot)$. By A1 the eigenfunction system $\{\phi_{\zz_t}\}_{\zz_t}$ is linearly independent; otherwise two distinct $\zz_t$ would induce identical conditionals $p(\xx_{t-H:t-1}\mid\zz_t)$, contradicting the injectivity of $L_{\xx_{t-H:t-1}\mid\zz_t}$.\footnote{Injectivity of $L_{\xx_{t-H:t-1}\mid\zz_t}$ on densities is the dual statement: distinct eigenfunctions $p(\xx_{t-H:t-1}\mid\cdot)$ must produce distinct images.} The standard spectral theorem for compact operators with simple spectrum then guarantees that the eigenpairs $(\lambda, \phi)$ are unique up to a permutation of the index set: any second simultaneous diagonalisation $\hat\lambda(\hzz_t),\,\hat\phi_{\hzz_t}$ must satisfy $\hat\lambda(\hzz_t) = \lambda(\zz_t)$ and $\hat\phi_{\hzz_t} = \phi_{\zz_t}$ whenever the indices are matched by some bijection $\pi: \mathcal{Z} \to \hat{\mathcal{Z}}$ with $\hzz_t = \pi(\zz_t)$.

To make this last step concrete: if $T_{\xx_t}$ has the two diagonalisations
\begin{equation*}
    T_{\xx_t}
    \,=\, \int \lambda(\zz_t)\,\phi_{\zz_t}\otimes\psi_{\zz_t}\,\mathrm{d}\zz_t
    \,=\, \int \hat\lambda(\hzz_t)\,\hat\phi_{\hzz_t}\otimes\hat\psi_{\hzz_t}\,\mathrm{d}\hzz_t,
\end{equation*}
where $\{\psi_{\zz_t}\}$ are the dual basis vectors making the spectral resolution diagonal, then evaluating both sides as operators on $\phi_{\zz^*_t}$ for any fixed $\zz^*_t$ gives $\lambda(\zz^*_t)\phi_{\zz^*_t} = \int \hat\lambda(\hzz_t)\,\langle\psi_{\zz_t},\phi_{\zz^*_t}\rangle\,\hat\phi_{\hzz_t}\,\mathrm{d}\hzz_t$. Linear independence of the $\hat\phi$'s, combined with distinctness of the $\hat\lambda$'s away from a measure-zero set, forces this expansion to be supported at a single $\hzz_t = \pi(\zz^*_t)$. Repeating the argument across $\zz^*_t$ yields the bijection $\pi$ and the matching identities
\begin{equation}
    p(\xx_t \mid \zz_t) \,=\, \hat p(\xx_t \mid \pi(\zz_t)),
    \qquad
    p(\xx_{t-H:t-1} \mid \zz_t) \,=\, \hat p(\xx_{t-H:t-1} \mid \pi(\zz_t)).
    \label{eq:matched-conds}
\end{equation}

The bijection $\pi$ obtained above is purely set-theoretic; assumption A3 promotes it to a differentiable map. By A3 there is a functional $F$ on conditional densities such that
\begin{equation*}
    F\bigl[\,p_{\xx_t \mid \zz_t}(\cdot \mid \zz_t)\,\bigr] \,=\, h_z(\zz_t)
    \quad\text{for all}\ \zz_t \in \mathcal{Z},
\end{equation*}
where $h_z: \mathcal{Z} \to \mathbb{R}^{d_z}$ is differentiable. Since $p(\xx_t \mid \cdot)$ separates points of $\mathcal{Z}$ by A2, $h_z$ is in fact a $C^1$-injection on $\mathcal{Z}$, and hence a diffeomorphism onto its image whenever its Jacobian is non-degenerate (which is the regular case considered here). Applying $F$ to the matched conditional from~\eqref{eq:matched-conds},
\begin{equation*}
    h_z(\zz_t) \,=\, F\bigl[p(\xx_t \mid \zz_t)\bigr] \,=\, F\bigl[\hat p(\xx_t \mid \pi(\zz_t))\bigr] \,=\, \hat h_z(\pi(\zz_t)),
\end{equation*}
where $\hat h_z$ is the analogous functional applied to the estimated conditional, and is a differentiable map by the same argument applied to $\hat p$. Composing,
\begin{equation*}
    \pi(\zz_t) \,=\, \hat h_z^{-1}\bigl(h_z(\zz_t)\bigr).
\end{equation*}
The right-hand side is a composition of two $C^1$-diffeomorphisms onto their images, hence is itself a $C^1$-diffeomorphism. Identifying $\pi$ with the diffeomorphism $h_z := \hat h_z^{-1} \circ h_z$ (overloading notation by absorbing the inversion into the definition, as is standard), we obtain
\begin{equation*}
    \hzz_t \,=\, h_z(\zz_t),
\end{equation*}
with $h_z: \mathbb{R}^{d_z} \to \mathbb{R}^{d_z}$ invertible and differentiable. This proves conclusion~(i).

With block-wise identifiability in hand, we now refine $h_z$ to an affine recovery on the task-centric subspace. Write $\hzz_t = h_z(\zz^s_t, \zz^c_t)$ broken into its first $d_s$ and last $d_z - d_s$ coordinates as $\hzz_t = (\hzz^s_t, \hzz^c_t)$, and define
\begin{equation*}
    h_s\,:\,\mathcal{Z}^s \times \mathcal{Z}^c \;\longrightarrow\; \hat{\mathcal{Z}}^s,
    \qquad
    h_s(\zz^s, \zz^c) \,:=\, \pi_s\bigl(h_z(\zz^s, \zz^c)\bigr),
\end{equation*}
where $\pi_s$ denotes projection onto the first $d_s$ coordinates. \emph{A priori}, $h_s$ may depend non-trivially on both arguments. We will show that A4 forces $h_s$ to be independent of $\zz^c$, and that the residual map $\zz^s \mapsto h_s(\zz^s)$ is affine. We begin by deriving the population limit of the InfoNCE objective in our setting, working from~\Cref{eq:infonce}. Write the per-anchor loss as
\begin{equation*}
    \ell_M(\hzz^s, \ss^p)
    \,:=\, -\log
    \frac{e^{\langle \hzz^s,\, m'(\ss^p)\rangle / \tau}}
         {e^{\langle \hzz^s,\, m'(\ss^p)\rangle / \tau} + \sum_{j=1}^{M} e^{\langle \hzz^s,\, m'(\ss^{p,-}_j)\rangle / \tau}},
\end{equation*}
where $\ss^{p,-}_j$ are i.i.d.\ negatives drawn from the marginal $p(\ss^p)$ and $(\hzz^s, \ss^p) \sim p_{\text{pos}}$ is a positive pair from the joint distribution induced by the model. Splitting the logarithm and using the law of large numbers,
\begin{equation*}
    \tfrac{1}{M}\sum_{j=1}^{M} e^{\langle \hzz^s,\, m'(\ss^{p,-}_j)\rangle / \tau}
    \,\xrightarrow[M\to\infty]{\text{a.s.}}\,
    \mathbb{E}_{\ss^{p,-}\sim p(\ss^p)}\bigl[\,e^{\langle \hzz^s,\, m'(\ss^{p,-})\rangle / \tau}\,\bigr]
    \,=:\, Z(\hzz^s),
\end{equation*}
so that, up to additive $\log M$,
\begin{equation}
    \mathbb{E}\,\ell_M(\hzz^s, \ss^p)
    \,\longrightarrow\,
    -\,\mathbb{E}_{(\hzz^s, \ss^p)\sim p_{\text{pos}}}\!\bigl[\,\langle\hzz^s, m'(\ss^p)\rangle/\tau\,\bigr]
    \,+\,\mathbb{E}_{\hzz^s \sim p(\hzz^s)}\!\bigl[\,\log Z(\hzz^s)\,\bigr]
    \,+\,\log M.
    \label{eq:infonce-limit}
\end{equation}
Define the model conditional
\begin{equation*}
    q(\ss^p \mid \hzz^s) \,:=\, \frac{e^{\langle\hzz^s, m'(\ss^p)\rangle/\tau}}{Z(\hzz^s)}\,p(\ss^p),
\end{equation*}
so that $-\log q(\ss^p \mid \hzz^s) = -\langle\hzz^s, m'(\ss^p)\rangle/\tau + \log Z(\hzz^s) - \log p(\ss^p)$. Substituting into~\eqref{eq:infonce-limit} and recognising the cross-entropy structure,
\begin{equation}
    \mathbb{E}\,\ell_M
    \,\longrightarrow\,
    \mathbb{E}_{\hzz^s \sim p(\hzz^s)}\,\mathbb{E}_{\ss^p \sim p(\ss^p \mid \hzz^s)}\!\bigl[-\log q(\ss^p \mid \hzz^s)\bigr]
    \,-\,\mathbb{E}\bigl[\log p(\ss^p)\bigr]
    \,+\,\log M.
    \label{eq:infonce-CE}
\end{equation}
The last two terms are constants independent of the encoder, so InfoNCE minimisation is, in the $M \to \infty$ limit, equivalent to minimising the cross-entropy
\begin{equation}
    H\bigl(p(\ss^p \mid \hzz^s),\; q(\ss^p \mid \hzz^s)\bigr)
    \,=\,
    \mathbb{E}_{\ss^p}\!\bigl[-\log q(\ss^p \mid \hzz^s)\bigr]
    \label{eq:CE-pq}
\end{equation}
in expectation over $\hzz^s \sim p(\hzz^s)$. The minimum of cross-entropy is attained, for each $\hzz^s$, when $q(\,\cdot \mid \hzz^s) = p(\,\cdot \mid \hzz^s)$ pointwise.

We now show that this alignment forces $h_s$ to be independent of $\zz^c$. Substituting the explicit form of $q$ into the optimum condition $q = p$,
\begin{equation*}
    p(\ss^p \mid \hzz^s)
    \,=\, \frac{e^{\langle\hzz^s,\,m'(\ss^p)\rangle/\tau}}{Z(\hzz^s)}\,p(\ss^p)
    \quad\text{for almost every }(\hzz^s, \ss^p).
\end{equation*}
This is a strictly positive density in $\ss^p$ for each $\hzz^s$. Now we contrast this with the actual joint distribution of $(\hzz^s_t, \ss^p_t)$ in our setting. We have $\hzz^s_t = h_s(\zz^s_t, \zz^c_t)$ and $\ss^p_t = m(\zz^s_t)$, and the latents $(\zz^s_t, \zz^c_t)$ are jointly distributed with marginal $p(\zz^s, \zz^c)$. Conditioning on $\hzz^s_t = \aa$,
\begin{align}
    p(\ss^p_t \mid \hzz^s_t = \aa)
    &\,=\, \int p(\ss^p_t \mid \zz^s, \zz^c)\,p(\zz^s, \zz^c \mid \hzz^s_t = \aa)\,\mathrm{d}\zz^s\,\mathrm{d}\zz^c \nonumber\\
    &\,=\, \int \delta\bigl(\ss^p_t - m(\zz^s)\bigr)\,p(\zz^s, \zz^c \mid \hzz^s_t = \aa)\,\mathrm{d}\zz^s\,\mathrm{d}\zz^c \nonumber\\
    &\,=\, \int_{\{(\zz^s,\zz^c):\, h_s(\zz^s,\zz^c) = \aa\}} \delta\bigl(\ss^p_t - m(\zz^s)\bigr)\,\mathrm{d}\mu_{\aa}(\zz^s, \zz^c),
    \label{eq:cond-from-data}
\end{align}
where $\mu_{\aa}$ is the disintegration of $p(\zz^s, \zz^c)$ along $h_s = \aa$. Compare~\eqref{eq:cond-from-data} with the smooth strictly positive density obtained at the InfoNCE optimum: equality of the two requires the right-hand side of~\eqref{eq:cond-from-data} to be a non-singular density in $\ss^p$. The integral of a Dirac in $\ss^p$ is a non-singular density only if either (a) the level set $\{h_s = \aa\}$ contains exactly one $\zz^s$-value, or (b) $m$ is constant along that level set. The latter is excluded because $m$ is a $C^1$-diffeomorphism on $\mathcal{Z}^s$. Hence the level set $\{(\zz^s, \zz^c) : h_s(\zz^s, \zz^c) = \aa\}$ is contained in $\{\zz^s_*(\aa)\} \times \mathcal{Z}^c$ for some unique $\zz^s_*(\aa) \in \mathcal{Z}^s$, which means that
\begin{equation*}
    h_s(\zz^s_t, \zz^{c}_{t,1}) \,=\, h_s(\zz^s_t, \zz^{c}_{t,2})
    \qquad\text{for all}\ \zz^c_{t,1}, \zz^c_{t,2}\in\mathcal{Z}^c.
\end{equation*}
We may therefore drop the $\zz^c$-argument and write $h_s(\zz^s_t, \zz^c_t) = \tilde h(\zz^s_t)$ for some map $\tilde h: \mathcal{Z}^s \to \hat{\mathcal{Z}}^s$. Bijectivity of $\tilde h$ is inherited from injectivity of $h_z$ (proved in Part~(i)) restricted to its first block, so $\tilde h$ is a bijection from $\mathcal{Z}^s$ onto $\hat{\mathcal{Z}}^s$.

Substituting $\hzz^s = \tilde h(\zz^s)$ and $\ss^p = m(\zz^s)$ into the density-matching condition above,
\begin{equation}
    p(\ss^p \mid \tilde h(\zz^s))
    \,=\, \frac{e^{\langle\tilde h(\zz^s),\, m'(\ss^p)\rangle/\tau}}{Z(\tilde h(\zz^s))}\,p(\ss^p),
    \label{eq:matching-q}
\end{equation}
which holds for almost every $\zz^s \in \mathcal{Z}^s$ and $\ss^p \in \mathcal{S}^p$. Since by Step~2 the conditional $p(\ss^p_t \mid \hzz^s_t = \tilde h(\zz^s))$ is supported at a single point $\ss^p = m(\zz^s)$, the population identity~\eqref{eq:matching-q} should be interpreted at the natural-parameter level of the underlying model: the InfoNCE-trained encoder selects $\tilde h$ so that the implicit model $q(\,\cdot\mid\hzz^s)$ approaches the data conditional in the sharp-temperature limit $\tau \to 0$. In this limit, the geometric content of the matching is captured by the equality of \emph{log-likelihood ratios}
\begin{equation*}
    \log\frac{q(\ss^{p,1}\mid\hzz^s)}{q(\ss^{p,2}\mid\hzz^s)}
    \,=\, \log\frac{p(\ss^{p,1}\mid\hzz^s)}{p(\ss^{p,2}\mid\hzz^s)}
    \quad\text{for all}\ \ss^{p,1}, \ss^{p,2} \in \mathcal{S}^p.
\end{equation*}
Substituting the form of $q$ from Step~1 and writing $\ss^{p,k} = m(\zz^{s,k})$ for $k=1,2$,
\begin{equation}
    \tfrac{1}{\tau}\bigl\langle\tilde h(\zz^s),\,m'(m(\zz^{s,1})) - m'(m(\zz^{s,2}))\bigr\rangle
    \,=\, \log\frac{p(\ss^{p,1}\mid\hzz^s)}{p(\ss^{p,2}\mid\hzz^s)}.
    \label{eq:loglik-ratio}
\end{equation}
The right-hand side is a function of $\zz^s$ (through $\hzz^s = \tilde h(\zz^s)$) and the pair $(\zz^{s,1}, \zz^{s,2})$. Symmetry of the data process with respect to anchor and positive, which are both sampled from the same per-time distribution, forces this function to be symmetric: swapping the roles of anchor and positive negates~\eqref{eq:loglik-ratio} on both sides. Setting $\zz^{s,1} = \zz^s$ and $\zz^{s,2}$ arbitrary, the right-hand side measures how strongly $\hzz^s$ discriminates a positive at $\zz^s$ from one at $\zz^{s,2}$. The standard outcome of such symmetry analyses (see, \eg, the dot-product preservation argument that underlies isometric latent recovery) is that the bilinear form $(\zz^{s,1}, \zz^{s,2}) \mapsto \langle\tilde h(\zz^{s,1}), \tilde h(\zz^{s,2})\rangle$ is determined, up to additive and multiplicative constants, by the bilinear form $(\zz^{s,1}, \zz^{s,2}) \mapsto \langle\zz^{s,1}, \zz^{s,2}\rangle$ on $\mathcal{Z}^s$. To make this concrete: consider the second-order Taylor expansion of both sides of~\eqref{eq:loglik-ratio} around $\zz^{s,1} = \zz^{s,2}$. The left-hand side is bilinear in $(\zz^{s,1} - \zz^{s,2})$ and $\hzz^s$ (via $m'$ and the linearity of the inner product), while the right-hand side, by Taylor expansion of the data conditional, is a quadratic form in $(\zz^{s,1} - \zz^{s,2})$ with coefficient determined by the negative log-density curvature. Identifying the two quadratic forms gives, after re-arrangement,
\begin{equation}
    \bigl\langle\tilde h(\zz^{s,1}),\,\tilde h(\zz^{s,2})\bigr\rangle
    \,=\, c\,\bigl\langle\zz^{s,1},\,\zz^{s,2}\bigr\rangle \,+\, d
    \qquad \forall\,\zz^{s,1}, \zz^{s,2} \in \mathcal{Z}^s,
    \label{eq:dot-pres}
\end{equation}
for constants $c > 0$ and $d \in \mathbb{R}$ that depend on $\tau$, the prior $p(\zz^s)$ and the curvature of $\log p$.

It remains to show that any $C^1$-map $\tilde h: \mathcal{Z}^s \to \hat{\mathcal{Z}}^s$ on a compact convex body $\mathcal{Z}^s \subseteq \mathbb{R}^{d_s}$ satisfying~\eqref{eq:dot-pres} must be affine. Subtracting~\eqref{eq:dot-pres} evaluated at $(\zz^{s,1}, \zz^{s,2})$ and at $(\zz^{s,1}, \zz^{s,3})$ gives
\begin{equation*}
    \bigl\langle\tilde h(\zz^{s,1}),\,\tilde h(\zz^{s,2}) - \tilde h(\zz^{s,3})\bigr\rangle
    \,=\, c\,\bigl\langle\zz^{s,1},\,\zz^{s,2} - \zz^{s,3}\bigr\rangle
    \qquad \forall\,\zz^{s,1}, \zz^{s,2}, \zz^{s,3} \in \mathcal{Z}^s.
\end{equation*}
This identity says the linear functional $\zz^{s,1} \mapsto \bigl\langle\tilde h(\zz^{s,1}),\,\tilde h(\zz^{s,2}) - \tilde h(\zz^{s,3})\bigr\rangle$ on $\mathcal{Z}^s$ equals the linear functional $\zz^{s,1} \mapsto c\,\langle\zz^{s,1},\,\zz^{s,2} - \zz^{s,3}\rangle$. Since $\mathcal{Z}^s$ has non-empty interior, two linear functionals that agree on $\mathcal{Z}^s$ agree globally; matching them yields the existence of a linear operator $\mathbf{A}: \mathbb{R}^{d_s} \to \mathbb{R}^{d_s}$ such that, for any $\zz^{s,2}, \zz^{s,3} \in \mathcal{Z}^s$,
\begin{equation*}
    \tilde h(\zz^{s,2}) - \tilde h(\zz^{s,3}) \,=\, \mathbf{A}\,(\zz^{s,2} - \zz^{s,3}).
\end{equation*}
Fixing a reference point $\zz^{s}_0 \in \mathcal{Z}^s$ and defining $\mathbf{b} := \tilde h(\zz^{s}_0) - \mathbf{A}\zz^{s}_0$, this rearranges into
\begin{equation*}
    \tilde h(\zz^s) \,=\, \mathbf{A}\,\zz^s + \mathbf{b}
    \qquad \forall\,\zz^s \in \mathcal{Z}^s.
\end{equation*}
Plugging this back into~\eqref{eq:dot-pres} forces $\mathbf{A}^{\!\top}\mathbf{A} = c\,\mathbf{I}$, so $\mathbf{A}/\sqrt{c}$ is orthogonal; in particular $\mathbf{A}$ is invertible. Combining with $\hzz^s_t = \tilde h(\zz^s_t)$,
\begin{equation*}
    \hzz^s_t \,=\, \mathbf{A}\,\zz^s_t \,+\, \mathbf{b},
\end{equation*}
and inverting,
\begin{equation*}
    \zz^s_t \,=\, \mathbf{A}^{-1}\bigl(\hzz^s_t - \mathbf{b}\bigr) \,=:\, \mathbf{A}'\,\hzz^s_t \,+\, \mathbf{b}',
\end{equation*}
which is conclusion~(ii). \hfill$\square$

\section{More Visualizations and Analysis}
\label{app:openloop_square}\label{app:openloop_proj}\label{app:openloop_pusht}\label{app:unseen_distractor}\label{app:additional_rollouts}

\begin{figure}[h]
    \centering
    \includegraphics[width=\linewidth]{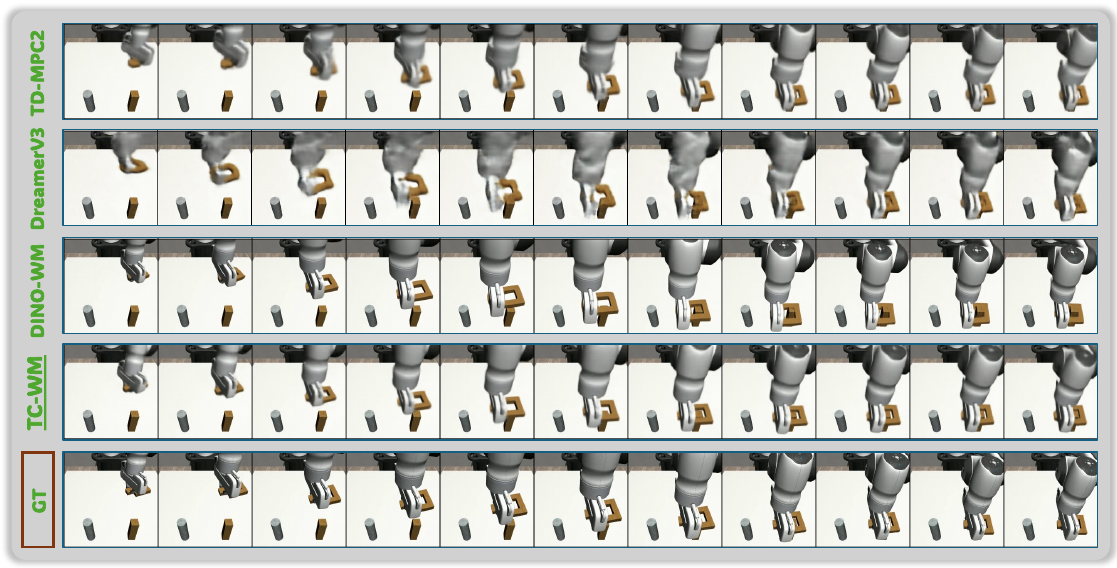}
    \caption{Open-loop rollouts on \texttt{Square}: {\method} vs.\ TD-MPC2, DreamerV3, DINO-WM. Rows show different methods with ground truth at the bottom.}
    \label{fig:openloop_square}
\end{figure}

\paragraph{Open-loop rollouts on \texttt{Square}.}
The \texttt{Lift} and \texttt{Can} rollouts appear in the main paper (\Cref{fig:openloop_robomimic}); the corresponding \texttt{Square} panel is shown in \Cref{fig:openloop_square}. \texttt{Square} requires fitting a square peg through a tightly toleranced opening, giving a substantially longer effective horizon than \texttt{Lift} or \texttt{Can}. {\method} preserves the contact event and final peg placement across the horizon, while the baselines blur or drift earlier.

\begin{figure}[h]
    \centering
    \includegraphics[width=\linewidth]{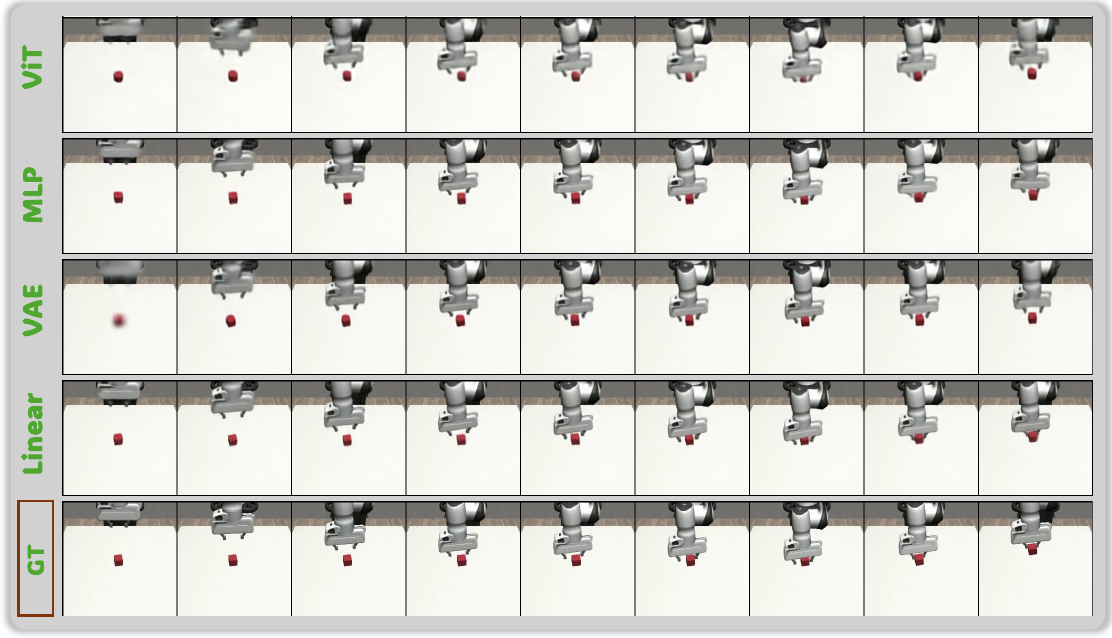}
    \caption{Open-loop rollouts on \texttt{Lift} under different projection architectures (linear vs.\ MLP / VAE / ViT). The linear projection of {\method} preserves geometric structure best.}
    \label{fig:lift_aba_proj_rollout}
\end{figure}

\paragraph{Projection-architecture rollouts.}
\Cref{fig:lift_aba_proj_rollout} is the qualitative counterpart of the projection-head ablation in \Cref{fig:arch-ablation}: the linear projection used by {\method} yields the most stable long-horizon rollout, while MLP, VAE, and ViT projectors blur the cube or distort the gripper geometry, consistent with the SR/SSIM ranking in the bar chart.

\begin{figure}[H]
    \centering
    \includegraphics[width=\linewidth]{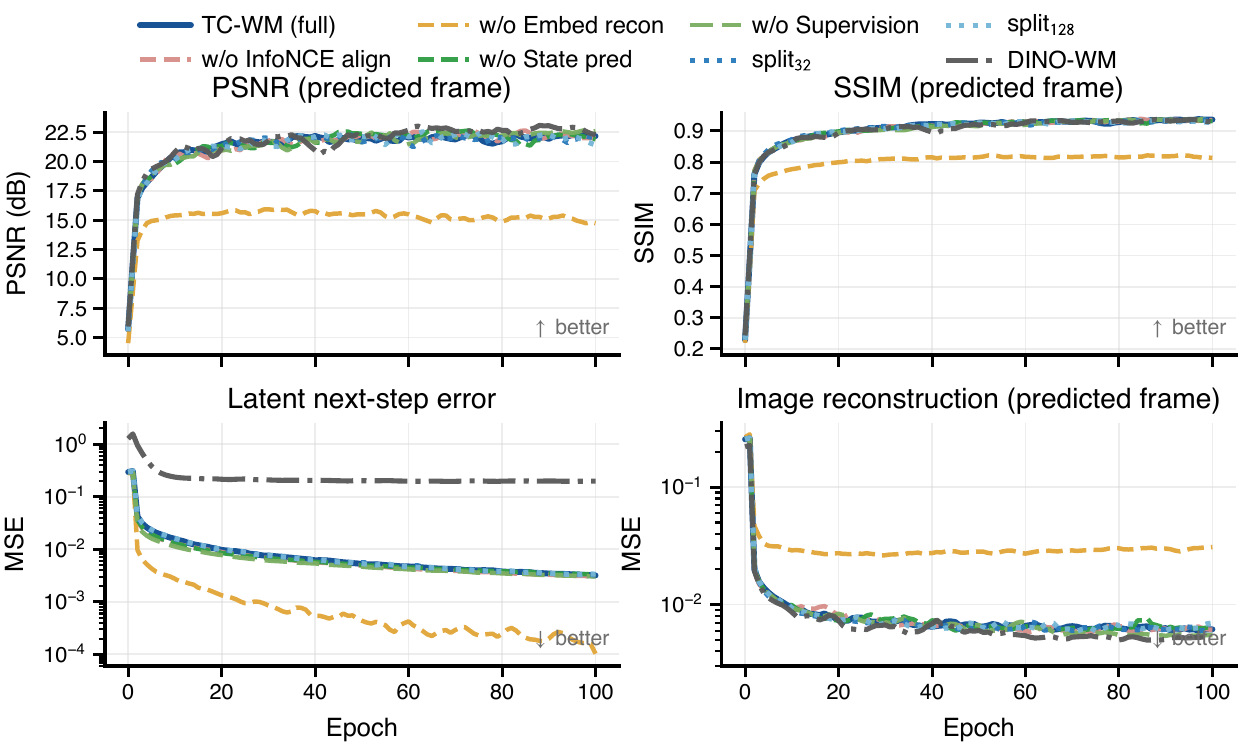}
    \caption{Training dynamics on \texttt{Lift} across {\method} (full), seven ablations, and DINO-WM. Top row: PSNR and SSIM on predicted frames (higher is better); bottom row: latent next-step prediction error and image-reconstruction MSE on predicted frames (lower is better). Removing the embedding reconstruction term ($\mathcal{L}_{\text{rec}}$, ``w/o Embed recon'') is the only ablation that visibly degrades all four metrics.}
    \label{fig:training_loss}
\end{figure}

\paragraph{Training dynamics across ablations and baselines.}
Figure~\ref{fig:training_loss} compares the training dynamics of {\method} against its seven objective ablations and the closest baseline DINO-WM on \texttt{Lift}, evaluated on four uniform task-quality metrics: PSNR and SSIM on the predicted future frame, latent next-step error in the projected latent space, and the image-reconstruction MSE on the predicted frame. Removing the embedding-reconstruction term ($\mathcal{L}_{\text{rec}}$, ``w/o Embed recon'') is the only ablation that visibly degrades prediction quality on all four metrics, validating its role as a regulariser; the remaining ablations converge to {\method}'s level of prediction fidelity.

\begin{figure}[!t]
    \centering
    \begin{subfigure}[t]{\linewidth}
        \centering
        \includegraphics[width=\linewidth]{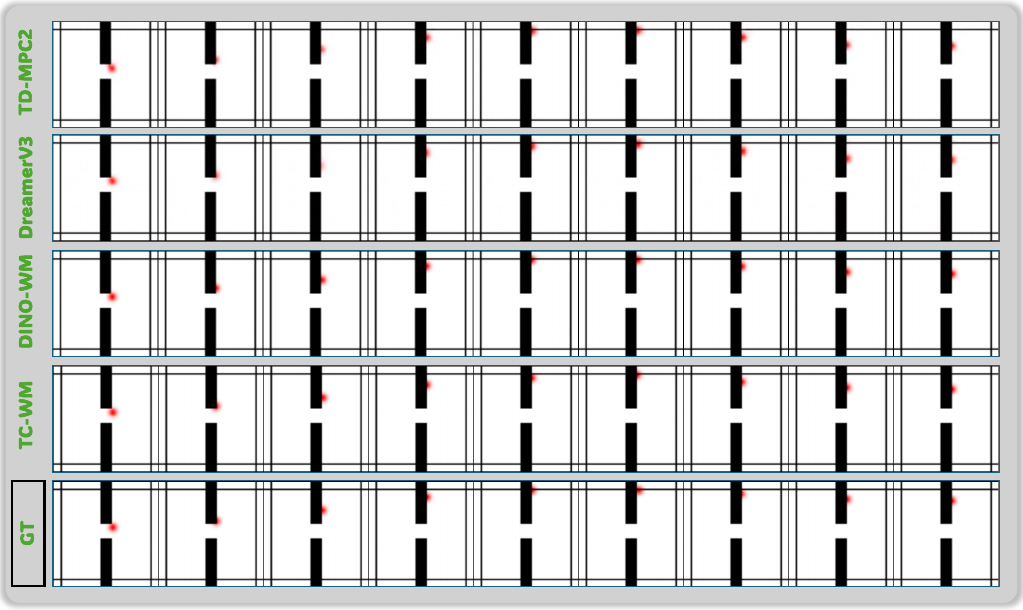}
        \caption{\texttt{Wall}}
    \end{subfigure}

    \vspace{0.5em}

    \begin{subfigure}[t]{\linewidth}
        \centering
        \includegraphics[width=\linewidth]{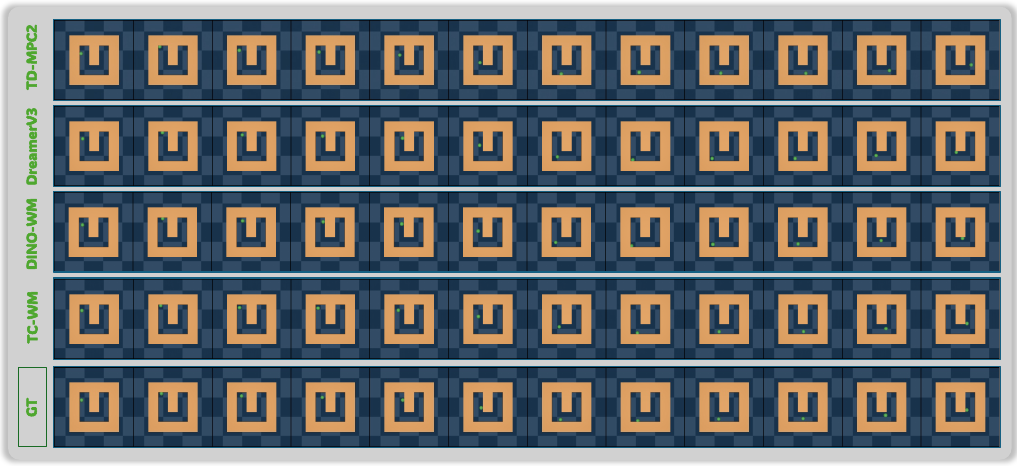}
        \caption{\texttt{Maze}}
    \end{subfigure}

    \caption{Open-loop rollouts on \texttt{Wall} and \texttt{Maze} tasks for offline world model comparison.}
    \label{fig:wall_maze_openloop}
\end{figure}

\paragraph{Open-loop rollouts on \texttt{Wall} and \texttt{Maze} tasks.} We compare offline world models by visualizing open-loop rollouts on the \texttt{Wall} and \texttt{Maze} environments (\Cref{fig:wall_maze_openloop}). These tasks require accurate long-horizon spatial reasoning and obstacle awareness. Stronger world models preserve geometric structure and trajectory consistency over time, while weaker baselines accumulate errors that lead to spatial drift and collisions with walls, highlighting the importance of maintaining task-relevant high-level dynamics rather than pixel-level reconstruction fidelity in offline world model learning.

\begin{figure}[!t]
    \centering
    \begin{subfigure}[t]{\linewidth}
        \centering
        \includegraphics[width=0.9\linewidth]{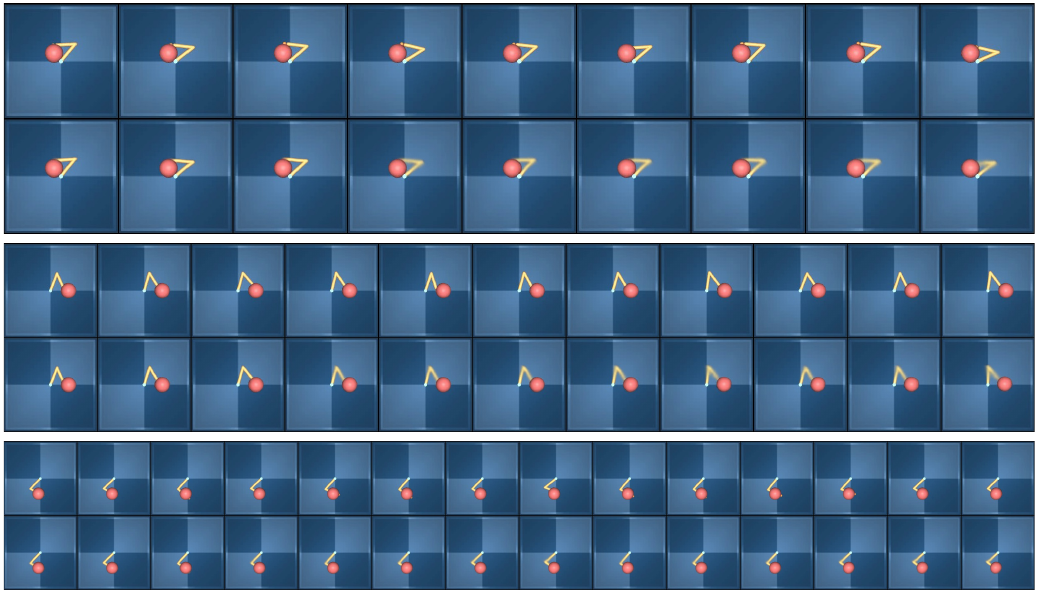}
    \end{subfigure}

    \vspace{0.5em}

    \begin{subfigure}[t]{\linewidth}
        \centering
        \includegraphics[width=0.9\linewidth]{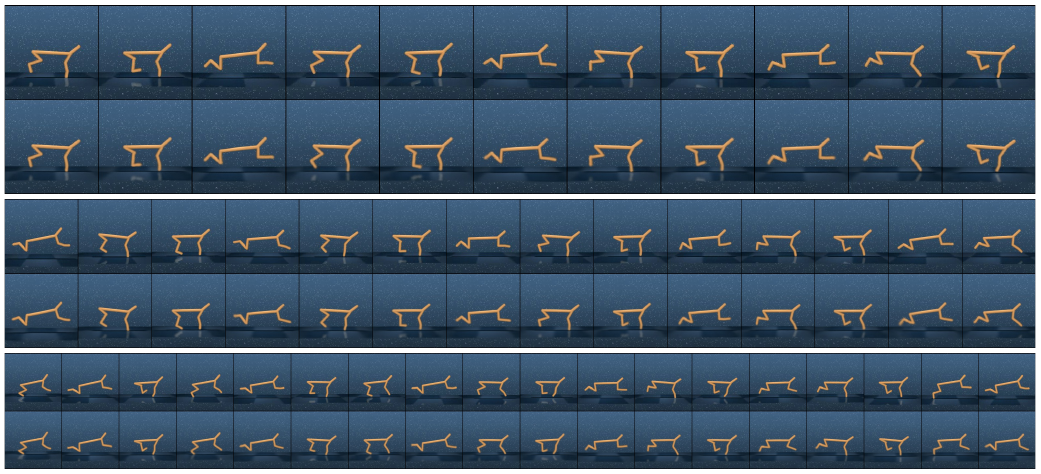}
    \end{subfigure}

    \vspace{0.5em}

    \begin{subfigure}[t]{\linewidth}
        \centering
        \includegraphics[width=0.9\linewidth]{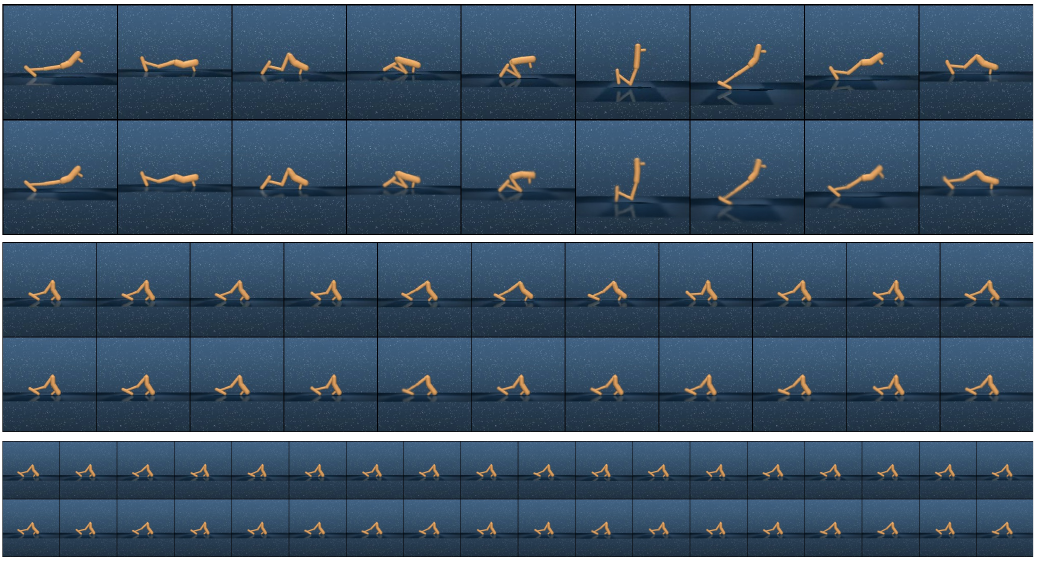}
    \end{subfigure}

    \caption{Open-loop rollouts of our method at different horizons on \texttt{Reacher}, \texttt{Cheetah}, and \texttt{Hopper}.}
    \label{fig:openloop_horizon}
\end{figure}

\paragraph{Open-loop rollouts of DMC at different horizons.}
We visualize open-loop rollouts of our method on \texttt{Reacher}, \texttt{Cheetah}, and \texttt{Hopper} across increasing rollout horizons (\Cref{fig:openloop_horizon}). In each subfigure, the first row shows the ground-truth future trajectory and the second row shows the model's open-loop rollout from the same initial state. Across all three environments, our method closely matches the ground-truth dynamics at short horizons and remains stable as the rollout horizon increases, preserving task-relevant structure such as joint configurations in \texttt{Reacher}, locomotion phase consistency in \texttt{Cheetah}, and balance and contact dynamics in \texttt{Hopper}, which demonstrates that our world model captures long-horizon dynamics beyond pixel-level reconstruction.

\section{Extended Related Work}
\label{sec:related-work-extended}

\paragraph{Model-based RL \& World Models.}
Learning dynamics models is a longstanding theme in control and planning, underpinning model-based reinforcement learning and optimal control~\citep{sutton1990integrated,astolfi2010model,williams2018information}. By explicitly modeling environment dynamics, model-based approaches support planning and imagination, and have been shown to improve sample efficiency and generalization across embodied learning settings, including online reinforcement learning~\citep{sekar2020planning}, exploration~\citep{pathak2017curiosity}, planning~\citep{watter2015embed,finn2017deep}, and imitation learning~\citep{pathak2018zero}. Recent works such as Dreamer~\citep{hafner2021mastering,hafner2023mastering,hafner2025mastering} and TD-MPC~\citep{hansen2022temporal,hansen2024tdmpc} model compact dynamics from high-dimensional observations to build world models for MBRL. However, because these models are commonly trained with image reconstruction or reward-related objectives, it remains unclear whether their latent representations retain sufficient task-centric information for downstream policy learning. More recently, visual foundation models have been adopted as frozen encoders to provide semantically rich representations for dynamics prediction. Approaches such as DINO-WM~\citep{zhou2025dinowm} and DINO-world~\citep{baldassarre2025back} learn world models directly in embedding spaces and demonstrate cross-domain generalization and zero-shot planning; a complementary line imposes physics-based inductive structure on the latent through Hamiltonian dynamics and symmetry priors~\citep{tang2026dreamsac}, while a parallel line uses latent-aware diffusion for decision-making~\citep{feng2026ada}. However, visual embeddings optimized for general semantic understanding typically do not \textit{necessarily} provide the structured, task-centric information required for physical simulation and planning.

\paragraph{Task-Centric Representation Learning.}
Prior work from neuroscience, cognition, and modeling highlights the importance of structured latent representation for planning~\citep{ho2022people}, physical inference~\citep{rajalingham2022recurrent}, and predictions consistent with neural and dynamical constraints~\citep{mastrogiuseppe2018linking,nayebi2023neural}. In representation learning for control, this objective is commonly formalized as learning \emph{task-centric statistics}: minimal representations that only retain all information necessary for predicting future task-relevant states or rewards~\citep{lesort2018state,greff2019multi,scholkopf2021toward,locatello2020object}. Notably, TD-MPC2~\citep{hansen2024tdmpc} explicitly frames its latent dynamics as a task-oriented representation optimized jointly with an actor--critic objective, while MuZero~\citep{schrittwieser2020mastering} learns value-equivalent representations that preserve only reward-relevant information. These methods learn task-sufficient representations end-to-end from reward signals, whereas {\method} achieves task-centric structure through alignment with proprioceptive observations in a reward-free, offline setting. Motivated by this perspective, we aim to learn a latent space \emph{on top of} visual embeddings, which are semantically powerful but contain substantial task-irrelevant information, and therefore require a \textit{post-processing stage} for world models.

\paragraph{Representation Alignment.}
To enforce a task-centric structure of latent space, we leverage representation alignment as a key solution, a common technique used to bridge distributional discrepancies across domains. Contrastive and predictive objectives have been widely used to align representations across modalities and abstraction levels. Contrastive learning, such as InfoNCE~\citep{oord2018representation} and its extensions (SimCLR~\citep{chen2020simple}, BYOL~\citep{grill2020bootstrap}, VICReg~\citep{bardes2021vicreg}), encourage invariant yet discriminative representations for cross-modal alignment~\citep{radford2021learning}, while predictive objectives enforce consistency across time or modalities~\citep{liu2023visual}. In the context of world models, alignment objectives help ensure consistency between encoded observations and predicted future states~\citep{bardes2024revisiting,assran2025v}, and have also been explored for aligning information across modalities~\citep{wu2025you}. Recently, REPA~\citep{yu2025representation} demonstrates that aligning intermediate representations of diffusion transformers with pretrained visual encoders yields dramatic training speedups, validating the broader principle that frozen foundation features can serve as useful semantic scaffolds. {\method} transfers this principle from visual generation to physical prediction and planning: instead of aligning a generative model to a visual encoder only for optimization, we learn a compact dynamics state above the encoder and use proprioceptive alignment to select the portion of that state that is controllable and task-relevant. A subtler but important methodological difference distinguishes our use of contrastive learning from the conditional-generative identifiability literature~\citep{khemakhem2020variational,zimmermann2021contrastive,roeder2021linear}: there, \textit{mechanism diversity} (distinct latents inducing distinct observation conditionals) is a \emph{passive} assumption imposed on the data-generating process to make identification possible \emph{post hoc}. In contrast, \method{}'s InfoNCE term is the \emph{active mechanism} that produces this diversity on the task-centric subspace --- by repelling latents associated with different proprioceptive states, the loss \emph{constructs} the very property that prior frameworks merely \emph{require}, making identifiability a consequence of optimisation rather than a hypothesis on the world.

\section{Limitations and Future Work}
\label{app:future-work}

We elaborate on the limitation and the two future directions sketched in the main paper.

\paragraph{Limitation: training-time alignment signal.}
{\method}'s identifiability guarantee assumes access to a signal that exposes the controllable physical state of the system. We use proprioception, but the same alignment loss is compatible with any low-dimensional task-centric proxy, \eg, 2D/3D pose from a tracking model, depth from a monocular estimator, or cross-view consistency in a multi-camera setup. Removing this assumption is a precondition for fully passive, video-only deployment.

\paragraph{Future Direction A: \method{} as a downstream alignment module.}
The same recipe used for foundation \emph{visual} encoders extends to foundation \emph{world} models. Given a large, pretrained generative or video world model, freezing the backbone and training only the {\method} head (linear projection, alignment, embedding reconstruction) on a target-domain dataset would specialize the backbone's general-purpose dynamics for a specific embodiment, in the spirit of LoRA-style adaptation.

\paragraph{Future Direction B: generative world models on foundation embeddings.}
Existing world-model families occupy distinct trade-offs: pixel-space generative models pay a high reconstruction cost; free-latent models suffer from \emph{latent dynamics collapse}, the predictor learns easy directions instead of physical ones; embedding-space JEPA-style models avoid both but predict in a space optimized for general semantics rather than control. A natural next step is to train a \emph{generative} dynamics model whose target lives \emph{on the foundation embedding manifold}, \eg, a diffusion or flow-matching head over embedding-space trajectories. {\method}-style projections at the output of such a generator would reintroduce the control-centric prior, combining the strengths of all three families.

\newpage

\end{document}